\documentclass[10pt,twocolumn,letterpaper]{article}

\usepackage{authblk}

\usepackage[pagenumbers]{cvpr} % To force page numbers, e.g. for an arXiv version

\definecolor{cvprblue}{rgb}{0.21,0.49,0.74}
\usepackage[pagebackref,breaklinks,colorlinks,allcolors=cvprblue]{hyperref}
\usepackage{multirow}
\usepackage{makecell}
\usepackage{float}

\usepackage[misc]{ifsym}

\newcommand{\name}{HoVLE}

\title{HoVLE: Unleashing the Power of Monolithic Vision-Language Models with \\Holistic Vision-Language Embedding}

\author[ ]{\vspace{-1em}Chenxin Tao$^{1,3}$\thanks{Equal contribution. \\ \indent \Letter\ Corresponding Author:  daijifeng@tsinghua.edu.cn.}}
\author[1,3*]{Shiqian Su}
\author[1,2*]{Xizhou Zhu}
\author[1,2]{Chenyu Zhang}
\author[4,2]{\\Zhe Chen}
\author[6]{Jiawen Liu}
\author[5,2]{Wenhai Wang}
\author[3]{Lewei Lu}
\author[1]{Gao Huang}
\author[2]{Yu Qiao}
\author[1,2 \Letter]{Jifeng Dai}
\affil[1]{Tsinghua University \quad $^2$Shanghai Artificial Intelligence Laboratory}
\affil[3]{SenseTime Research \quad $^4$Nanjing University}
\affil[5]{The Chinese University of Hong Kong \quad $^6$
Johns Hopkins University\vspace{-2em}}

\begin{document}
\maketitle
\begin{minipage}{\textwidth}
\center
\vspace{-3em}
    \noindent Model: \url{https://huggingface.co/OpenGVLab/HoVLE}
\end{minipage}
\begin{abstract}

The rapid advance of Large Language Models (LLMs) has catalyzed the development of Vision-Language Models (VLMs). Monolithic VLMs, which avoid modality-specific encoders, offer a promising alternative to the compositional ones but face the challenge of inferior performance. Most existing monolithic VLMs require tuning pre-trained LLMs to acquire vision abilities, which may degrade their language capabilities. To address this dilemma, this paper presents a novel high-performance monolithic VLM named \name{}. We note that LLMs have been shown capable of interpreting images, when image embeddings are aligned with text embeddings. The challenge for current monolithic VLMs actually lies in the lack of a holistic embedding module for both vision and language inputs. Therefore, \name{} introduces a holistic embedding module that converts visual and textual inputs into a shared space, allowing LLMs to process images in the same way as texts. Furthermore, a multi-stage training strategy is carefully designed to empower the holistic embedding module. It is first trained to distill visual features from a pre-trained vision encoder and text embeddings from the LLM, enabling large-scale training with unpaired random images and text tokens. The whole model further undergoes next-token prediction on multi-modal data to align the embeddings. Finally, an instruction-tuning stage is incorporated. Our experiments show that \name{} achieves performance close to leading compositional models on various benchmarks, outperforming previous monolithic models by a large margin.

\end{abstract}
    
\begin{figure}[ht]
    \centering
    \includegraphics[width=\linewidth]{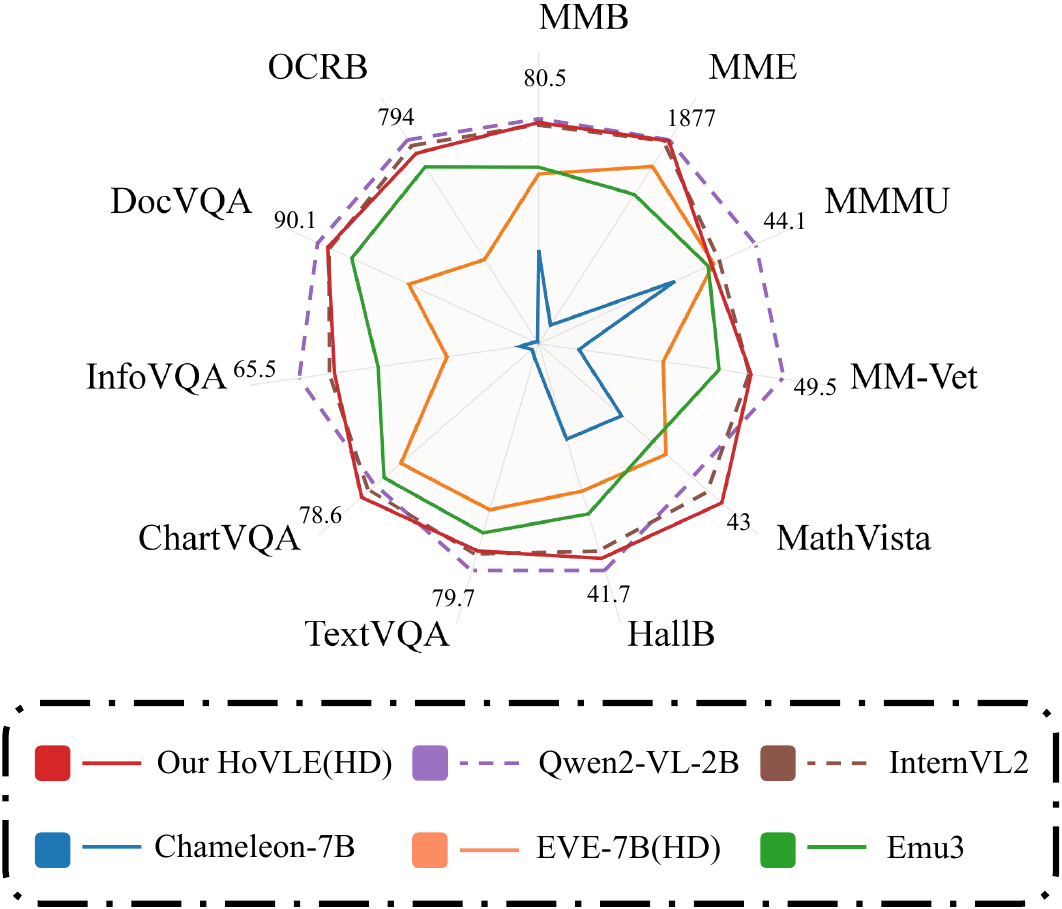}
    \vspace{-1.5em}
    \caption{\textbf{Performance comparison on different benchmarks between compositional VLMs (dashed lines) and monolithic VLMs (solid lines).} 
    Previous monolithic VLMs exhibit a significant performance gap compared to compositional VLMs, 
    while Our \name{} demonstrate competitive capabilities with state-of-the-art compositional VLMs.}
    \vspace{-1.5em}
    \label{fig:teaser_radar}   
\end{figure}

\section{Introduction}
\label{sec:intro}

\begin{figure}[ht]
    \centering
    \includegraphics[width=\linewidth]{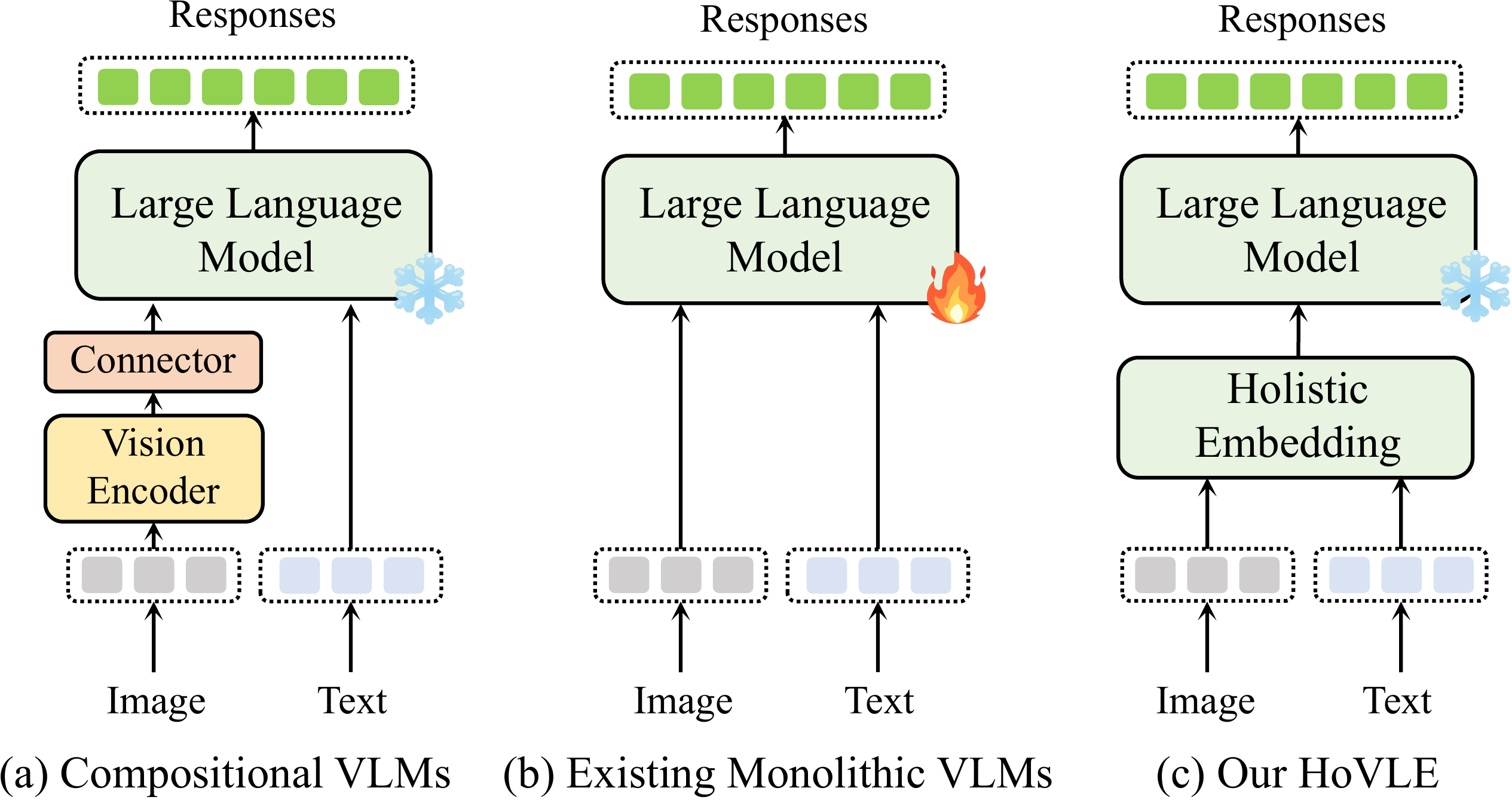}
    \vspace{-1.5em}
    \caption{\textbf{Comparison of VLM architectures.} (a) \textit{Compositional VLMs} integrate pre-trained vision encoders with LLMs, using an extra connector to align image and text embeddings. (b) Existing \textit{Monolithic VLMs} directly feed image and text inputs into LLMs, which require continual pre-training to gain visual abilities. (c) \textit{\name{}} uses a holistic embedding module to project image and text input to a unified embedding space, enabling LLMs to interpret images in a text-like manner. Blocks with the same color have the same Transformer layer architecture.}
    \vspace{-1em}
    \label{fig:teaser}
\end{figure}

The rapid advancement of Large Language Models (LLMs)~\cite{zhao2023survey, touvron2023llama, achiam2023gpt, bai2023qwen, cai2024internlm2} has significantly accelerated the development of Vision-Language Models (VLMs)~\cite{liu2024improved,wang2024visionllm,zhu2023minigpt}, which have demonstrated powerful visual-text processing capabilities. 
\emph{Compositional VLMs} are currently the \textit{de facto} approach for building high-performance VLMs~\cite{chen2024internvl,wang2024qwen2,liu2024llavanext}, which integrate pre-trained vision encoders with LLMs, as shown in Fig.~\ref{fig:teaser}(a). By combining pre-trained visual understanding capabilities with language modeling capabilities, compositional VLMs achieve impressive multi-modal performance.
On the other hand, there is growing interest in developing VLMs without modality-specific encoders, referred to as \emph{monolithic VLMs} in this paper. These models process image and text inputs as a unified whole, directly feeding them into the LLM, as illustrated in Fig.~\ref{fig:teaser}(b). 
Due to their monolithic design, these VLMs feature simpler architectures, offering greater potential for multimodal tasks, such as unifying generation and recognition~\cite{team2024chameleon,wang2024emu3,zhou2024transfusion}. Monolithic VLMs are gaining increasing attention and have shown their potential across many studies~\cite{fuyu-8b,diao2024unveiling,chen2024single}.

However, monolithic VLMs still lag behind compositional VLMs in terms of performance. Some researches~\cite{fuyu-8b,team2024chameleon,zhou2024transfusion,beyer2024paligemma} explore training monolithic VLMs from scratch using a combination of multi-modal and pure text data. Such practice fails to leverage existing pre-trained models, resulting in substantial computational demands for training. Another research direction~\cite{diao2024unveiling,chen2024single} proposes injecting visual understanding capabilities into pre-trained LLMs by conducting continual pre-training~\cite{gururangan2020don} with large-scale and high-quality multi-modal data. Nevertheless, such continual-tuning methods may interfere with the language knowledge already acquired by pre-trained LLMs, potentially limiting their multi-modal performance~\cite{diao2024unveiling}.

The performance limitations of current monolithic VLMs mainly stem from the need to tune pre-trained LLMs. However, the success of compositional VLMs indicates that pre-trained LLMs can interpret visual information without further pre-training, as long as image embeddings are properly aligned with text embeddings. Thus, the reason for such LLM-tuning dilemma of current monolithic VLMs is the lack of a holistic embedding module that can coordinate different modalities.

To this end, we develop a modality-sharing holistic embedding module as shown in Fig.~\ref{fig:teaser}(c), which maps mixed visual and language inputs into a shared embedding space. It can enable pre-trained LLMs to interpret visual input in the same way as text. Similar to compositional VLMs, this not only avoids tuning pre-trained LLMs, but also preserves their language proficiency while successfully integrating visual capabilities. Our experiments demonstrate that such a holistic embedding module can indeed achieve high-performance monolithic VLMs.

In this work, we introduce \name{}, a novel monolithic vision-language model that learns a \textbf{Ho}listic \textbf{V}ision-\textbf{L}anguage \textbf{E}mbedding module to extend LLMs with vision capabilities. The holistic embedding module consists of the same causal Transformer layers as the LLM. It accepts both images and texts as input, and projects them into a unified embedding space. These embeddings are then forwarded into the LLM, constituting a monolithic VLM. 
To empower the holistic embedding module with vision and language encoding abilities, a multi-stage training strategy is proposed. First, the module is trained to distill visual features from a pre-trained vision encoder and text embeddings from the LLM. The distillation stage does not require image-text pairs, but instead uses large-scale unpaired random images and text tokens as inputs.
After distillation, the holistic embedding module is combined with the LLM to perform next token prediction on multi-modal data, so that the output embeddings can be further aligned into a unified space. Finally, the whole model is enhanced with an instruction tuning stage. The alignment and instruction tuning stages are similar to those used in compositional VLMs, eliminating the need for extra supervised data.

Extensive experiments show that our \name{} is able to achieve close performance with these leading compositional VLMs on 17 multi-modal benchmarks. It also surpasses previous monolithic VLMs by a large margin, e.g., $\sim$15 points on MMBench~\cite{liu2025mmbench}, showing the effectiveness of \name{}.
In conclusion, our contributions can be summarized as follows:
\begin{itemize}
    \item \name{} is proposed as a novel monolithic VLM that learns a holistic embedding module to expand LLMs with vision capabilities. By projecting vision and language inputs to a shared embedding space, \name{} can understand images without compromising language proficiency.
    \item A novel multi-stage training strategy is proposed to empower the holistic embedding module with vision and language encoding abilities. Our training procedure enables \name{} to develop strong vision and language capabilities without requiring additional supervised data compared to compositional VLMs.   
    \item \name{} demonstrates comparable performance with compositional VLMs, and surpasses previous monolithic VLMs by a large margin.
\end{itemize}

\section{Related Work}
\label{sec:related}

\textbf{Compositional Vision-Language Models.}
The rapid advancements and success of Large Language Models (LLMs) have sparked researchers' interest in multi-modal models, with Vision-Language Models (VLMs) emerging as a prominent focus~\cite{chen2024internvl, diao2024unveiling, li2022blip, liu2024improved, sun2024generative, liu2024visual}. Compositional VLMs have become the dominant paradigm in the field, where image tokens are typically generated through visual encoders and an additional connector (e.g., MLP projection). Many notable commercial models, including GPT-4V~\cite{yang2023dawn}, the Gemini series~\cite{team2023gemini}, Claude-3V~\cite{TheC3}, Grok-1.5V~\cite{grok}, and MM1~\cite{mckinzie2024mm1}, have demonstrated impressive performance across various multi-modal tasks. In the open-source domain, models like LLaVA~\cite{liu2024visual, liu2024improved}, BLIP~\cite{li2022blip}, Emu2~\cite{sun2024generative}, InternVL~\cite{chen2024internvl}, and Qwen-VL~\cite{bai2023qwen} stand out. However, compositional VLMs face several challenges, such as complexities in deployment and optimization due to their heterogeneous architecture, trade-offs in the design of network capacity between image and text processing, and potential limitations in cross-modal understanding capabilities~\cite{diao2024unveiling, chen2024single,team2024chameleon, beyer2024paligemma}.

\vspace{0.5em}
\noindent\textbf{Monolithic Vision-Language Models.}
In contrast to compositional VLMs, monolithic VLMs do not rely on a specific vision encoder. 
Some works explore training a monolithic VLM from scratch. Fuyu-8B~\cite{fuyu-8b} directly inputs image pixel patches into the backbone network using a simple linear projection, while Chameleon~\cite{team2024chameleon}, Transfusion~\cite{zhou2024transfusion}, and Emu3~\cite{wang2024emu3} adopt image tokenizers to map images to discrete tokens as inputs. Nevertheless, these methods require collecting massive pure text and paired image-text data, as well as intensive training computational resources to achieve good performance. 
Other approaches propose to extend existing pre-trained LLMs to VLMs. SOLO~\cite{chen2024single} uses a simple linear projection to handle raw image pixels and fine-tunes the pre-trained LLM to develop vision abilities. EVE~\cite{diao2024unveiling} employs a patch embedding layer with cross-attention to enhance image representation, alongside a patch aligning layer to provide fine-grained visual supervision with pre-trained vision encoders. However, these methods require continually pre-training LLMs, which may diminish their language proficiency. Current monolithic VLMs still underperform in image-text understanding tasks when compared to their compositional counterparts. 

In contrast, by introducing a holistic embedding module, \name{} can achieve comparable performance with state-of-the-art compositional VLMs, significantly surpassing other monolithic VLMs.

\vspace{0.5em}
\noindent\textbf{Knowledge Distillation.}
Knowledge distillation is an effective technique for transferring knowledge and abilities from a teacher model to a student model. In the realm of VLMs, knowledge distillation has also been successfully incorporated into various tasks. DistillVLM~\cite{fang2021compressing} and EfficientVLM~\cite{wang2022efficientvlm} use knowledge distillation to compress the model size and accelerate inference speed. VPD~\cite{hu2024visual} and LSKD~\cite{park2024localized} leverage tools or LLMs to synthesize multi-modal data so that VLMs can be trained to distill the skills in specialist models and LLMs. 
Our work employs knowledge distillation to endow the holistic embedding module with general vision and language encoding abilities. Since distillation targets can be computed independently, we can use unpaired random image and text data, eliminating the need to collect additional multi-modal data.

\section{Method}
\label{sec:method}

\begin{figure*}
    \centering
    \includegraphics[width=0.95\linewidth]{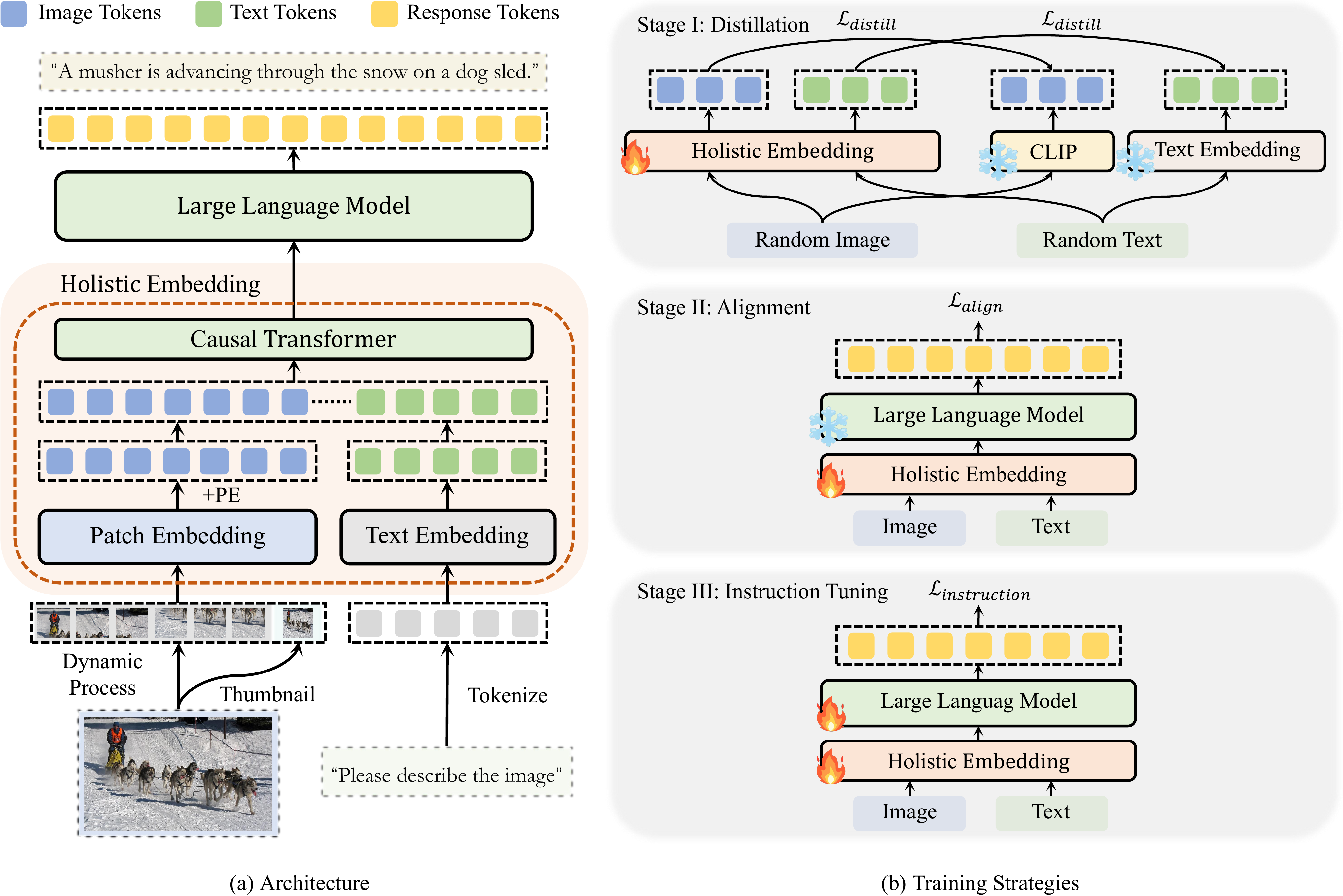}
    \vspace{-0.5em}
    \caption{
    \textbf{(a) The architecture of \name{}.}
    \name{} initially segments the input images into patches dynamically and tokenizes input texts. The holistic embedding module then projects them into a unified space. Finally, the LLM processes these unified embeddings to produce the final outputs.
    \textbf{(b) The training strategies of \name{}.}
    \textit{Distillation stage} trains the holistic embedding to distill a pre-trained vision encoder and text embeddings of the LLM using unpaired random images and texts.
    \textit{Alignment stage} combines the holistic embedding module with a frozen LLM, conducting auto-regressive training to align the vision-language embeddings.
    \textit{Instruction tuning} further enhance \name{}'s overall ability by tuning the whole model.
    }
    \label{fig:overview}
    \vspace{-1.5em}
\end{figure*}

In this section, we present \name{}, a monolithic vision-language model with decent vision and language capabilities. In the following sections, we introduce the network architecture and the training procedures towards such a unified model. The details are shown in Figure~\ref{fig:overview}.

\subsection{Network Architecture}
The success of compositional VLMs demonstrates that LLMs are capable of understanding visual information when image embeddings are properly aligned with text embeddings. Inspired by this, \name{} proposes that a monolithic VLM can be built using a holistic embedding module. This module takes mixed raw images and texts as inputs and projects them into a shared embedding space. In this way, LLMs can interpret images in the same way as texts, resulting in a monolithic VLM.

\vspace{0.5em}
\noindent\textbf{Holistic Embedding Module.} We first introduce the holistic embedding module to project vision and language inputs to a shared embedding space. Specifically, given an input image $I$ and texts $T$, we first transform them into vector forms via simple tokenization and linear projection:
\begin{align}
    x_I &= \text{PatchEmbed}(\text{DynProcess}(I)) + \text{PE}, \\
    x_T &= \text{TextEmbed}(\text{Tokenizer}(T)),
\end{align}
where $\text{DynProcess}(\cdot)$ is the dynamic high-resolution strategy from InternVL~\cite{chen2024internvl}. This process divides the image into $448\times 448$ tiles and appends a global thumbnail to provide overall context. Similar operations have been adopted in many compositional and monolithic VLMs~\cite{xu2024llava,huang2024hires,diao2024unveiling} and have proved beneficial for model performance. $\text{PatchEmbed}(\cdot)$ refers to a patch embedding layer with stride $s$ and converts the image tiles into a sequence $x_I$ of $n_I$ image tokens with dimension $c$. $\text{PE}\in \mathbb{R}^{n_I\times c}$ is a learnable position embedding.
$\text{Tokenizer}(\cdot)$ converts raw texts into token indices and $\text{TextEmbed}(\cdot)$ is a word embedding layer with vocabulary size of $v$ and dimension of $c$. $x_T$ is a sequence of $n_T$ text tokens.

Afterward, these image tokens and text tokens are combined as the input tokens $x\in \mathbb{R}^{n\times c}$. The holistic embedding module utilizes a stack of causal Transformer layers to project these tokens to a shared embedding space:
\begin{equation}
    \hat{x} = \text{CausalTransformer}(x),
\end{equation}
where $\hat{x}$ denotes final embeddings of the input.
The architecture of each Transformer layer is the same as the LLM's for simplicity and unity. Similar to LLMs, these Transformer layers use causal attention for both image and text tokens. The effectiveness of this architecture suggests that 1D causal modeling works as well as non-causal modeling for images in multi-modal tasks, aligning with recent findings in \cite{tao2024learning}.

\vspace{0.5em}
\noindent\textbf{Monolithic Vision-Language Model.} By combining the holistic embedding module with a pre-trained LLM, a monolithic VLM can be naturally built. After the holistic embedding module generates input embeddings, the LLM takes them as input and performs next-token prediction:
\begin{equation}
    p_{n+1} = \text{LLM}(\hat{x}; \theta_{\text{LLM}}),
\end{equation}
where $\theta_{\text{LLM}}$ refers to parameters of LLM, $p_{n+1}\in \mathbb{R}^{v}$ is the probability of the next token.

In this way, we replace the modality-specific encoders in compositional VLMs with a holistic embedding module. This module has the same Transformer layer with the LLM, so the whole model presents a simple architecture, leading to our monolithic VLM.

\subsection{Training Procedure}
To empower the holistic embedding module with decent vision and language encoding abilities, a multi-stage training procedure is carefully designed. The first stage trains the holistic embedding module to distill the image feature from a pre-trained visual encoder and the text embeddings from an LLM, providing general encoding abilities. The second stage combines the holistic embedding module with the LLM to perform auto-regressive training, aligning different modalities to a shared embedding space.

\vspace{0.5em}
\noindent\textbf{Distillation Stage.} This stage aims to provide basic vision and language encoding abilities with the holistic embedding module. We extract image and text embeddings from the holistic embedding module output, denoted as $\hat{x}_I$ and $\hat{x}_T$, respectively. A pre-trained visual encoder and an LLM embedding layer are utilized to compute the target image feature $z_I$ and text feature $z_T$. The training objective is
\begin{equation}
    \min_{\theta_{\text{Embed}}} \mathcal{L}_{\text{distill}}(\hat{x}_I, z_I; \theta_{\text{Embed}}) + \mathcal{L}_{\text{distill}}(\hat{x}_T, z_T; \theta_{\text{Embed}}),
\end{equation}
where $\mathcal{L}_{\text{distill}}$ is the negative cosine similarity between two features, $\theta_{\text{Embed}}$ refers to the parameters of the holistic embedding module.

\vspace{0.5em}
\noindent\textbf{Unpaired Random Images and Texts for Distillation.} The distillation stage does not put any requirement on input images and texts, because their target features are computed independently. Therefore, we propose to combine random images with random texts as input. In fact, this strategy proves to be more effective than using paired image-text inputs, potentially due to better vocabulary coverage during distillation, as demonstrated in Section~\ref{sucsec:ablation}. 

In practice, we randomly sample 500M images from LAION-2B~\cite{schuhmann2022laion} dataset, and randomly sample 100 text tokens from the vocabulary for each image as training data. Each sample consists of a user query and a model response. We place the image at the beginning of the query, followed by the text content for the query and response, each with 50 random text tokens. Unlike previous monolithic VLMs, \name{} can benefit from large-scale vision pre-training, unleashing the full potential of the monolithic VLM.

\vspace{0.5em}
\noindent\textbf{Alignment Stage.} To align the image and text embeddings, the holistic embedding module is further combined with the LLM to perform auto-regressive training. The training objective is
\begin{equation}
    \min_{\theta_{\text{Embed}}} \sum_{i}\mathcal{L}_{\text{align}}(p_{i}, y_i; \theta_{\text{Embed}}, \theta_{\text{LLM}}),
\end{equation}
where $\mathcal{L}_{\text{align}}$ is the cross-entropy loss, $p_i$ and $y_i$ denote the output probability and target index of the $i^{\text{th}}$ text token. During training, the LLM is frozen, and only the holistic embedding module is optimized.

This stage is similar to the alignment stage used in compositional VLMs, but we do not have an extra MLP connector. During alignment, the model is trained on 45M multi-modal data from diverse domains, such as captioning, detection, and OCR data. Please refer to Appendix~\ref{appendix:details} for details.

\subsection{Instruction Tuning}
The holistic embedding module is now empowered with strong vision and language encoding abilities. To further strengthen the whole VLM to follow instructions, a visual instruction tuning stage is incorporated.

In this stage, the whole model is optimized using the same auto-regressive training objective as in the alignment stage. The training data contains around 5M data, covering visual question answering, mathematics, world knowledge, etc. Please refer to Appendix~\ref{appendix:details} for details.

\section{Experiment}
\label{sec:exp}

\begin{table*}[t!]
\centering
\small
\setlength{\tabcolsep}{2.5pt}
\scalebox{1.0}{
\begin{tabular}{l c | c c c c c c c c c }
\toprule
\multirow{1}{*}{Method} & \multirow{1}{*}{\# Params} & \multicolumn{1}{c}{MMB} &  \multicolumn{1}{c}{MME} & MMMU  & \multirow{1}{*}{MM-Vet} &\multicolumn{1}{c} {POPE} & \multicolumn{1}{c} {SEED} & MathVista & HallB & CCB  \\
\midrule
\rowcolor{Gray!5}
\multicolumn{11}{l}{\textit{Compositional VLMs}} \\
VILA1.5-3B~\cite{lin2024vila}          & 3.1B  & 63.4 & 1648 & 33.3 & 35.4 & 85.9 & 67.9 & 31.6 & 31.2 & 24.1 \\ 
DeepSeek-VL-1.3B~\cite{lu2024deepseek}    & 2.0B  & 64.6 & 1532 & 32.2 & 34.8 & 87.6 & 66.7 & 31.1 & 27.6 & 37.6 \\
PaliGemma-3B~\cite{beyer2024paligemma}        & 2.9B  & 71.0 & 1686 & 34.9 & 33.1 & 87.0 & 69.6 & 28.7 & 32.2 & 29.6 \\
MM1-3B-Chat~\cite{mckinzie2024mm1}         & 3B    & 67.8 & 1762 & 33.9 & 43.7 & 87.4 & 68.8 & 32.0 & -    & -    \\
MiniCPM-V~\cite{yao2024minicpm}           & 2.8B  & 64.1 & 1650 & 38.3 & 31.1 & 79.5 & 65.6 & 28.9 & 36.2 & 41.4 \\ 
MiniCPM-V-2~\cite{yao2024minicpm}         & 2.8B  & 69.1 & 1809 & 38.2 & 41.0 & 86.3 & 67.1 & 38.7 & 36.1 & 45.3 \\ 
InternVL1.5~\cite{chen2024far}         & 2.2B  & 70.9 & 1902 & 34.6 & 39.3 & 87.1 & 69.8 & 41.1 & 37.5 & 63.5 \\ 
$^\dagger$InternVL2~\cite{internvl2}           & 2.2B  & 73.2 & 1877 & 34.3 & 44.6 & 88.3 & 71.6 & 46.4 & 37.9 & 74.7 \\ 
Qwen2-VL-2B~\cite{wang2024qwen2}         & 2.2B  & 74.9 & 1872 & 41.1 & 49.5 & -    & -    & 43.0 & 41.7 & -  \\
Phi-3 Vision~\cite{abdin2024phi}        & 4.2B  & 80.5 & -    & 40.4 & -    & 85.8 & -    & 44.5 & -    & -  \\
\midrule
\rowcolor{Gray!5}
\multicolumn{11}{l}{\textit{Monolithic VLMs}} \\
Fuyu-8B (HD)~\cite{fuyu-8b}        & 8B    & 10.7 & -    & -    & 21.4 & 74.1 & -    & -    & -   & -   \\
SOLO~\cite{chen2024single}                & 7B    & -    & 1260 & -    & -    & -    & 64.4 & 34.4 & -   & -     \\
$^*$Chameleon-7B~\cite{team2024chameleon}        & 7B    & 31.1 & 170  & 25.4 & 8.3  & - & 30.6 & 22.3  & 17.1 & 3.5   \\
EVE-7B~\cite{diao2024unveiling}              & 7B    & 49.5 & 1483 & 32.3 & 25.6 & 83.6 & 61.3 & 25.2 & 21.1 & 12.4 \\ 
EVE-7B (HD)~\cite{diao2024unveiling}         & 7B    & 52.3 & 1628 & 32.6 & 25.7 & 85.0 & 64.6 & 34.2 & 26.4 & 16.3 \\ 
Emu3~\cite{wang2024emu3}                & 8B    & 58.5 & -    & 31.6 & 37.2 & 85.2 & 68.2 & -    & -    & -   \\
\rowcolor{Gray!15}
Our \name{}     & 2.6B         & 71.9 & \textbf{1864} & \textbf{33.7} & \textbf{44.3} & \textbf{87.6} & 70.7 & 46.2 & \textbf{39.6} & \textbf{75.3} \\
\rowcolor{Gray!15}
Our \name{}~(HD) & 2.6B         & \textbf{73.3} & 1862 & 32.2 & 43.8 & 87.4 & \textbf{70.9} & \textbf{49.2} & 38.4 & 74.3 \\ 
\bottomrule
\end{tabular}
}
\vspace{-1em}
\caption{\textbf{Comparison with existing VLMs on general VLM benchmarks.} The highest scores in monolithic VLMs are highlighted in bold. $^*$We test Chameleon-7B with its released model, but it frequently refuses to perform the task with a response of “I can’t help you with this”, thus resulting in poor performance. $^\dagger$ InternVL2 2.2B adopts the same LLM and supervised data with our HoVLE, so we mark it as the compositional counterpart.}
\vspace{-1.5em}
\label{tab:results_general}
\end{table*}

\subsection{Implementation Details}

\noindent\textbf{Network Architecture.}
\name{} adopts InternLM2-1.8B~\cite{cai2024internlm2} as the LLM backbone. The holistic embedding is composed of 8 causal Transformer layers, with hidden size 2048 and 16 attention heads, identical to those in InternLM2-1.8B. The image patch embedding layer has a stride of $s=28$ and output dimension $c=2048$. The text embedding has a vocabulary of 92,553.
To demonstrate the high-resolution image processing ability of our model, like other works~\cite{diao2024unveiling,fuyu-8b,li2024monkey}, we also provide \name{}~(HD). It uses a patch embedding with stride $s=14$ and applies pixel shuffle~\cite{chen2024internvl} after the holistic embedding to reduce the number of visual tokens to 1/4. All other details are kept the same.

\vspace{0.5em}
\noindent\textbf{Training Procedure.}
During distillation, we randomly sample 500M images from LAION-2B~\cite{schuhmann2022laion} with 100 random text indices for each image. We use InternViT-300M~\cite{chen2024far,chen2024internvl} and the text embedding from InternLM2-1.8B as the teacher model for image and text tokens, respectively. 
For alignment, a collection of datasets containing 50M samples is used, as shown in Appendix~\ref{appendix:details}.
For instruction tuning, a collection of datasets containing 5M samples is used, as shown in Appendix~\ref{appendix:details}.
We use AdamW as the optimizer for all three stages, with learning rate 3e-4, 5e-5 and 4e-5, respectively. The batch size is set to 4096 for all stages. Please refer to Appendix~\ref{appendix:details} for more hyper-parameters.

\vspace{0.5em}
\noindent\textbf{Evaluation Benchmarks.}
We evaluate our model on 17 public benchmarks. MMbench~\cite{liu2025mmbench}, MME~\cite{fu2024mme}, MMMU~\cite{yue2024mmmu}, MM-Vet~\cite{yu2023mmvet}, and SEED~\cite{li2023seed} focus on multimodal perception and reasoning, providing a comprehensive assessment of VLM capabilities. POPE~\cite{li2023evaluating} and HallB~\cite{guan2023hallusionbench} specifically target hallucination detection. For OCR and OCR-related visual question answering, TextVQA~\cite{singh2019towards}, ChartQA~\cite{masry2022chartqa}, InfoVQA~\cite{mathew2022infographicvqa}, DocVQA~\cite{mathew2021docvqa}, and OCRBench~\cite{liu2023hidden} are used. GQA~\cite{hudson2019gqa} evaluates visual scene understanding, while ScienceQA~\cite{lu2022learn} and AI2D~\cite{kembhavi2016diagram} assess scientific image understanding. MathVista focuses on mathematical reasoning. For specific metric values shown in the experiments, MMbench-EN results are counted and presented as the MMB score. The POPE score is averaged across its three categories, the MME score is calculated as the sum of perception and cognition scores, and the SEED score reflects image accuracy specifically.

\subsection{Main Results}
We compare \name{} with previous compositional and monolithic VLMs in Table~\ref{tab:results_general} and Table~\ref{tab:results_vqa}.
We highlight several key observations as follows: 
(1) There are significant performance gaps between previous monolithic VLMs and compositional VLMs, indicating the challenges in building a monolithic VLM with state-of-the-art performance.
(2) Compared with existing monolithic VLMs, our \name{} demonstrates substantially better performance with fewer parameters. For example, \name{} surpasses the previous best monolithic VLM by $\sim$15 points on MMBench, which is a comprehensive multi-modal benchmark covering a variety of fine-grained abilities. This substantial improvement is attributed to the holistic embedding, which prevents continual pre-training from disrupting language knowledge stored in pre-trained LLMs.
(3) By incorporating higher resolution, \name{}~(HD) consistently improves the performance on VQA benchmarks, demonstrating the versatility of our holistic embedding in handling different resolutions.
(4) Compared with previous compositional VLMs, \name{}~(HD) is able to deliver competitive results with many of these models. 
This shows that the introduction of holistic embedding provides an effective encoding that allows the LLM to understand images while preserving its language abilities.
While \name{}~(HD) still falls short of state-of-the-art compositional VLMs, like Qwen2-VL-2B~\cite{wang2024qwen2} and Phi-3 Vision~\cite{abdin2024phi}, it narrows the gap between monolithic and compositional VLMs, suggesting the feasibility of a high-performance monolithic VLM.

\begin{table*}[t!]
\centering
\small
\setlength{\tabcolsep}{2.5pt}
\scalebox{0.95}{
\begin{tabular}{l c | c  c  c  c  c  c  c c }
\toprule
\multirow{1}{*}{Method} & \multirow{1}{*} {\# Params} &   \multirow{1}{*} {TextVQA} &  \multirow{1}{*} {ChartVQA} &  \multirow{1}{*} {InfoVQA} &  \multirow{1}{*} {DocVQA} & OCRBench & \multirow{1}{*} {GQA} & SQA-I & AI2D \\
\midrule
\rowcolor{Gray!5}
\multicolumn{10}{l}{\textit{Compositional VLMs}} \\
VILA1.5-3B~\cite{lin2024vila}          & 3.1B  & 60.4 & -    & - 
   & -    & 437 & 61.5 & 69.0 & -  \\
DeepSeek-VL-1.3B~\cite{lu2024deepseek}    & 2.0B  & 57.8 & -    & -    & -    & 409 & -     & -    & 51.5  \\
PaliGemma-3B~\cite{beyer2024paligemma}        & 2.9B  & 68.1 & -    & -    & -    & 614 & -     & -    & 68.3  \\
MM1-3B-Chat~\cite{mckinzie2024mm1}         & 3B    & 71.9 & -    & -    & -    & -    & -    & 69.4 & -    \\
MiniCPM-V~\cite{yao2024minicpm}           & 2.8B  & 60.6 & -    & -    & 38.2 & 366 & -     & -    & 56.3  \\
MiniCPM-V-2~\cite{yao2024minicpm}         & 2.8B  & 74.1 & -    & -    & 71.9 & 605  & -    & -    & 62.9  \\
InternVL1.5~\cite{chen2024far}         & 2.2B  & 70.5 & 74.8 & 55.4 & 85.0 & 654  & 61.6 & 84.9 & 69.8 \\
$^\dagger$InternVL2~\cite{internvl2}           & 2.2B  & 73.4 & 76.2 & 57.7 & 85.9 & 784  & 61.0 & 84.9 & 74.1 \\
Qwen2-VL-2B~\cite{wang2024qwen2}         & 2.2B  & 79.7 & 73.5 & 65.5 & 90.1 & 794  & -    & -    & -     \\
Phi3-Vision~\cite{abdin2024phi}         & 4.2B  & 70.9 & 81.4 & -    & -    & -    & -    & 90.8 & 76.7  \\
\midrule
\rowcolor{Gray!5}
\multicolumn{10}{l}{\textit{Monlithic VLMs}} \\
Fuyu-8B (HD)~\cite{fuyu-8b}        & 8B    & -    & -    & -    & -    & -    & -    & -    & 64.5 \\
SOLO~\cite{chen2024single}                & 7B    & -    & -    & -    & -    & -    & -    & 73.3 & 61.4 \\
$^*$Chameleon-7B~\cite{team2024chameleon}        & 7B    & 4.8  & 2.9  & 5.0  & 1.5  & 7  & -    & 47.2 & 46.0 \\
EVE-7B~\cite{diao2024unveiling}              & 7B    & 51.9 & 19.5 & 20.0 & 22.0 & 327 & 60.8 & 63.0 & 48.5 \\ 
EVE-7B (HD)~\cite{diao2024unveiling}         & 7B    & 56.8 & 59.1 & 25.0 & 53.0 & 398 & 62.6 & 64.9 & 59.1 \\
Emu3~\cite{wang2024emu3}                & 8B    & 64.7 & 68.6 & 43.8 & 76.3 & 687  & 60.3 & 89.2 & 70.0 \\ 
\rowcolor{Gray!15}
Our \name{}     & 2.6B  & 66.0 & 77.6 & 52.1 & 84.7 & 735 & \textbf{61.0} & \textbf{94.8} & \textbf{73.7} \\
\rowcolor{Gray!15}
Our \name{} (HD) & 2.6B & \textbf{70.9} & \textbf{78.6} & \textbf{55.7} & \textbf{86.1} & \textbf{740} & 60.4 & \textbf{94.8} & 73.0 \\ 
\bottomrule
\end{tabular}
}
\vspace{-1em}
\caption{\textbf{Comparison with existing VLMs on visual question answering benchmarks.} The highest scores in monolithic VLMs are highlighted in bold. 
Same as Tab.~\ref{tab:results_general}, $^*$Chameleon-7B frequently refuses to response, and $^\dagger$InternVL2 2.2B is our compositional counterpart.}
\vspace{-1em}
\label{tab:results_vqa}
\end{table*}

\begin{table}[]
    \centering
    \small
    \scalebox{0.95}{
    \setlength{\tabcolsep}{2pt}
    \begin{tabular}{c|c c c c c c}
    \toprule
         Depth & MMB & MME & SEED & TextVQA & InfoVQA & DocVQA \\
    \midrule
       0  & 35.4 & 1149 & 46.0 & 11.3 & 19.1 & 12.2 \\
       4  & 67.8 & 1801 & 68.0 & 47.7 & 42.6 & 73.6 \\
       8  & 72.0 & 1862 & 70.9 & 62.0 & 51.4 & 82.1 \\
       12 & 72.3 & 1882 & 70.9 & 65.5 & 52.5 & 84.7 \\
    \bottomrule
    \end{tabular}
    }
    \vspace{-1em}
    \caption{\textbf{Ablation on holistic embedding depth.}}
    \vspace{-1em}
    \label{tab:ablation_depth}
\end{table}
\begin{table}[]
    \centering
    \small
    \scalebox{0.95}{
    \setlength{\tabcolsep}{1pt}
    \begin{tabular}{c c c |c c c c c c}
    \toprule
         Dist. & Align. & I.T. & MMB & MME & SEED & TextVQA & InfoVQA & DocVQA \\
    \midrule
        & \checkmark & \checkmark & 37.3 & 1274 & 46.4 & 11.2 & 19.3 & 13.2 \\
        \checkmark & & \checkmark       & 71.3 & 1828 & 70.5 & 60.1 & 49.2 & 80.9 \\
        \checkmark & \checkmark &        & 66.1 & 1698 & 66.7 & 55.0 & 43.0 & 71.8 \\
       \checkmark & \checkmark & \checkmark & 72.0 & 1862 & 70.9 & 62.0 & 51.4 & 82.1 \\
    \bottomrule
    \end{tabular}
    }
    \vspace{-1em}
    \caption{\textbf{Ablation on training strategies.}}
    \vspace{-1em}
    \label{tab:training}
 \end{table}

\subsection{Ablation Studies}
\label{sucsec:ablation}
In this section, we aim to display and discuss the design choices that affect our \name{}. Ablation experiments train the model with 300M and 8M data during the distillation and alignment stages respectively, unless specified.

\vspace{0.5em}
\noindent\textbf{Holistic Embedding Depth.}
To validate the effectiveness of the holistic embedding, we first compare models with different holistic embedding depths in Table~\ref{tab:ablation_depth}. With 0 Transformer layer, the model degenerates to a VLM without holistic embedding and the poor performance suggests the importance of this module. The model performs consistently better as the depth goes from 4 to 12. We speculate that deep embedding is beneficial for extracting visual representation. \name{} uses 8 layers to achieve a balance between performance and model complexity.

\begin{figure*}
    \centering
    \includegraphics[width=\linewidth]{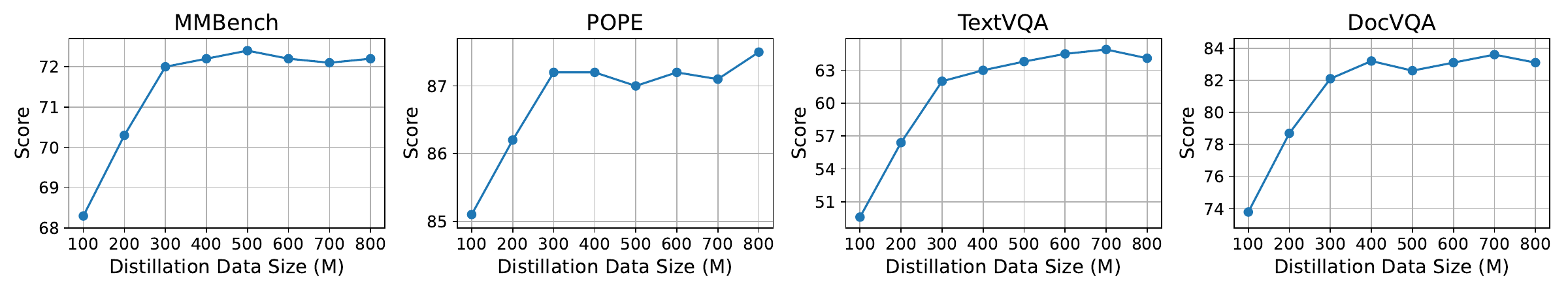}
    \vspace{-2em}
    \caption{\textbf{Distillation data scaling performance.}}
    \vspace{-1em}
    \label{fig:data_scaling}
\end{figure*}
\begin{table}[]
    \centering
    \small
    \scalebox{0.95}{
    \setlength{\tabcolsep}{2pt}
    \begin{tabular}{l |c c c c c c}
    \toprule
         Text Data & MMB & MME & SEED & TextVQA & InfoVQA & DocVQA \\
    \midrule
       Paired  & 71.7 & 1768 & 70.5 & 61.3 & 49.4 & 81.5 \\
       Random & 72.0 & 1862 & 70.9 & 62.0 & 51.4 & 82.1 \\
    \bottomrule
    \end{tabular}
    }
    \vspace{-1em}
    \caption{\textbf{Ablation on text data for distillation.}}
    \vspace{-1em}
    \label{tab:ablation_text_data}
\end{table}

\vspace{0.5em}
\noindent\textbf{Training Strategy.}
Table~\ref{tab:training} displays the influence of different training stages. The results show that omitting any stage will result in a drop of performance. Among these, removing the distillation stage leads to the most severe degradation, highlighting its importance. We observe that models after distillation exhibit a faster decrease in loss during subsequent training stages, likely because the distillation provides general abilities of representation extraction, which is helpful for accelerating training. The improvement from instruction tuning is greater than that of the alignment stage. We assume this is because the LLM has more trainable parameters, allowing the whole model to fit the data well.

\vspace{0.5em}
\noindent\textbf{Distillation Data Scaling.}
Figure~\ref{fig:data_scaling} illustrates how the performance changes as distillation data scales up. Overall, the model performs better with more distillation data, implying the importance of comprehensive training. 
Specifically, the model's performance improves rapidly with 300M data, and continues to increase at a more gradual pace as more data is added. We use 500M distillation data as our default setting. Please refer to Appendix~\ref{appendix:ablation} for scaling performance on more benchmarks.

\begin{figure*}[t]
    \centering
    \includegraphics[width=0.9\linewidth]{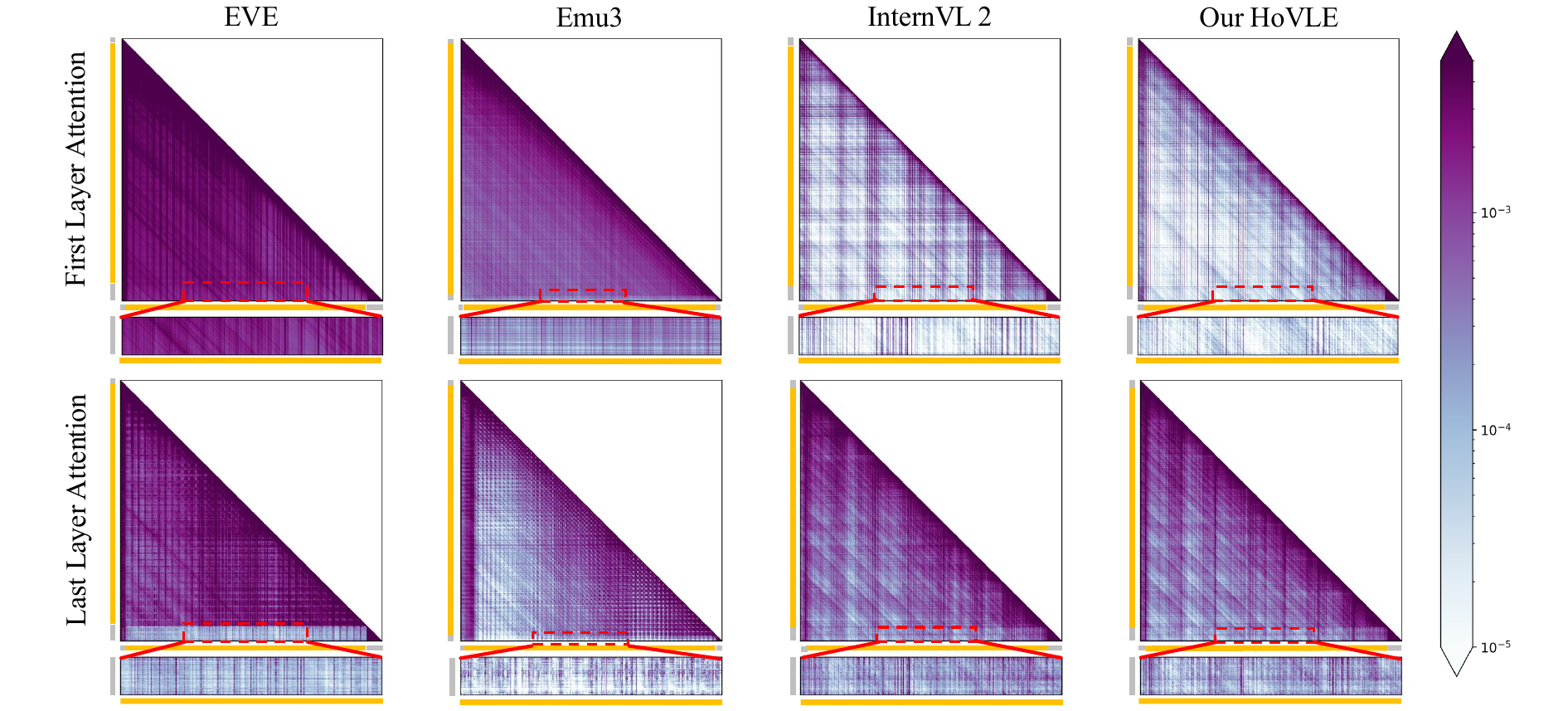}
    \vspace{-1em}
    \caption{\textbf{Attention Maps for EVE, Emu3, InternVL2 and our HoVLE at the first and last layers of LLM backbones}. Y-axis represents query tokens, and X-axis represents key tokens, with text modality tokens in gray and image modality tokens in yellow. All four models share the same input, but the sequence lengths of input tokens are different due to different image pre-processing. We highlight text-to-image attention below each full attention map. Our HoVLE, like the compositional InternVL2, has sparse attention across all network layers, while other monolithic models Emu3 and EVE have denser attention in shallow layers.}
    \vspace{-1.5em}
    \label{fig:vis}
\end{figure*}

\vspace{0.5em}
\noindent\textbf{Text Data for Distillation.}
Table~\ref{tab:ablation_text_data} presents the effects of different text input choices during distillation. The LAION~\cite{schuhmann2022laion} dataset provides a short text caption for each image, so we can also use image-text pairs as input. However, this approach yields relatively lower results, and the model continues to exhibit high loss values during alignment. We hypothesize that the text data from the LAION dataset lacks sufficient diversity to cover the full vocabulary, suggesting that paired data may require higher quality and variety in texts. In contrast, we propose using random texts as input. Table~\ref{tab:ablation_text_data} shows that this strategy outperforms paired text data. We also observe lower loss during alignment. By leveraging random text, we reduce the high-quality need for image-text pairs collection, thereby enhancing data utilization for large-scale training.

\subsection{Analysis}

\noindent\textbf{Visualization.}
To better understand the effects of our introduced holistic embedding, we visualize the attention map of the LLM in Figure~\ref{fig:vis}.
We find that existing monolithic models, such as EVE~\cite{diao2024unveiling} and Emu3~\cite{wang2024emu3}, exhibit denser text-to-image attention in their initial layers, while the compositional InternVL2~\cite{internvl2} and our HoVLE model maintain sparse attention patterns throughout. All models demonstrate sparse attention patterns in the later layers (with the final layers shown). Please refer to Appendix~\ref{appendix:vis} for more attention maps in the middle layers.

We hypothesize that sparser text-to-image attention indicates that visual features are effectively extracted and well-aligned with text tokens, allowing the text tokens to focus only on the most relevant visual information. In contrast, denser text-to-image attention may suggest that the model is still in the process of extracting visual features within the LLM component. Therefore, by introducing the holistic embedding, the LLM may process vision and language inputs more effectively.

\begin{table}[t!]
\centering
\small
\setlength{\tabcolsep}{2.5pt}
\scalebox{0.95}{
\begin{tabular}{l c | c c c c}
\toprule
Model & \makecell{Tile\\Resolution} &  FLOPs(T)$\downarrow$ &  TTFT$\downarrow$ &  TPS$\uparrow$ & Avg. Score$\uparrow$ \\
\midrule
InternVL2 & $448\times 448$ & 1.81 & 1.37 & 281 & 65.8 \\
\name{}~(HD) & $336\times 336$ & 1.51 & 1.24 & 243 & 65.5 \\
\name{}~(HD) & $392\times 392$ & 1.89 & 1.52 & 224 & 65.9 \\
\name{}~(HD) & $448\times 448$ & 2.18 & 1.84 & 206 & 66.5 \\
\bottomrule
\end{tabular}
}
\vspace{-0.5em}
\caption{\textbf{Inference speed-performance trade-off of \name{}~(HD).} “TTFT” and “TPS”
denotes the time to first token in seconds and throughput in tokens per second, respectively. "Avg. Score" refers to the average performance over 17 benchmarks. Please refer to Appendix for detailed results.}
\vspace{-2em}
\label{tab:speed}
\end{table}

\vspace{0.5em}
\noindent\textbf{Speed-Performance Trade-off.}
We report the speed-performance trade-off of \name{}~(HD) in Table~\ref{tab:speed}. For evaluation, we use one image and 256 text tokens as input, with the model restricted to output 120 tokens during testing. InternVL2 serves as the main comparison model, as it uses the same LLM. When the tile resolution of DynProcess($\cdot$) is 336, \name{}~(HD) achieves comparable performance to InternVL2, while also demonstrating lower computational demand. As the input resolution increases, \name{}'s performance improves consistently, but at the cost of slower speed. Additionally, we observe that using the KV cache~\cite{pope2023efficiently} during inference significantly reduces computational time, making kernel launch the main bottleneck in TPS. Optimizing the kernel could further enhance processing speed.
\section{Conclusion}
\label{sec:conclusion}
This work presents \name{}, a monolithic Vision-Language Model (VLM) that processes images and texts in a unified manner. \name{} introduces a holistic embedding module that projects image and text inputs into a shared embedding space. This allows the Large Language Model (LLM) to interpret images in the same way as texts. Furthermore, a multi-stage training procedure is designed to strengthen the encoding abilities of the holistic embedding. A distillation stage first provides general vision and language encoding capabilities, while an alignment stage aligns different modalities. An instruction tuning stage enhances the overall abilities in the end. Extensive experiments show that \name{} surpasses previous monolithic VLMs and demonstrates competitive performance with compositional VLMs. This work narrows the gap between monolithic and compositional VLMs, providing a promising direction for the development of monolithic VLMs.

\noindent\textbf{Limitations.} Due to limited computational resources, our experiments are conducted on 2B parameter scale. Scaling to larger scales can provide further insight into the model scalability of our approach, which we hope to explore in future work.

\noindent\textbf{Acknowledgements.} This work is supported by the National Key R\&D Program of China (NO. 2022ZD0161300), by the National Natural Science Foundation of China (62376134).
\clearpage
\appendix
\maketitlesupplementary

\section{Implementation Details}
\label{appendix:details}
\noindent\textbf{Hyper-parameters.} The hyper-parameters for three training stage of our \name{} are listed in Table~\ref{tab:appendix_distill_param}, Table~\ref{tab:appendix_align_param}, and Table~\ref{tab:appendix_sft_param}, respectively.

\vspace{0.5em}
\noindent\textbf{Datasets.} Table~\ref{tab:ds_align} and Table~\ref{tab:ds_sft} list the detailed datasets used in the alignment and instruction tuning stages.

\begin{table}[h]
    \centering
    \small
    \scalebox{0.95}{
    \setlength{\tabcolsep}{2pt}
    \begin{tabular}{c|c}
    \toprule
         Hyper-parameters & Value \\
    \midrule
       Resolution of Image tile  & $448\times 448$ \\
       Amount of data  & $500$M \\
       Batch size & 4096 \\
       Warmup steps & 2000 \\
       Optimizer & AdamW \\
       Peak learning rate & $3\times 10^{-4}$\\
       Learning rate schedule & Constant \\
       Weight decay & $0.05$\\
       AdamW $\beta$ & $(0.9, 0.999)$\\
       AdamW $\epsilon$ & $1\times 10^{-8}$\\
    \bottomrule
    \end{tabular}
    }
    \vspace{-1em}
    \caption{\textbf{Hyper-parameters for distillation stage}}
    \vspace{-1em}
    \label{tab:appendix_distill_param}
\end{table}
\begin{table}[h]
    \centering
    \small
    \scalebox{0.95}{
    \setlength{\tabcolsep}{2pt}
    \begin{tabular}{c|c}
    \toprule
         Hyper-parameters & Value \\
    \midrule
       Resolution of Image tile  & $448\times 448$ \\
       Amount of data  & $50$M \\
       Batch size & 4096 \\
       Warmup steps & 100 \\
       Optimizer & AdamW \\
       Peak learning rate & $5\times 10^{-5}$\\
       Learning rate schedule & Cosine \\
       Weight decay & $0.01$\\
       AdamW $\beta$ & $(0.9, 0.999)$\\
       AdamW $\epsilon$ & $1\times 10^{-8}$\\
    \bottomrule
    \end{tabular}
    }
    \vspace{-1em}
    \caption{\textbf{Hyper-parameters for alignment stage}}
    \vspace{-1em}
    \label{tab:appendix_align_param}
\end{table}
\begin{table}[h]
    \centering
    \small
    \scalebox{0.95}{
    \setlength{\tabcolsep}{2pt}
    \begin{tabular}{c|c}
    \toprule
         Hyper-parameters & Value \\
    \midrule
       Resolution of Image tile  & $448\times 448$ \\
       Amount of data  & $5$M \\
       Batch size & 4096 \\
       Warmup ratio & 0.03 \\
       Optimizer & AdamW \\
       Peak learning rate & $4\times 10^{-5}$\\
       Learning rate schedule & Cosine \\
       Weight decay & $0.01$\\
       AdamW $\beta$ & $(0.9, 0.999)$\\
       AdamW $\epsilon$ & $1\times 10^{-8}$\\
    \bottomrule
    \end{tabular}
    }
    \vspace{-1em}
    \caption{\textbf{Hyper-parameters for instruction tuning stage}}
    \vspace{-1em}
    \label{tab:appendix_sft_param}
\end{table}

\begin{table}[h]\scriptsize
\renewcommand{\arraystretch}{1.2}
\centering

    \begin{tabular}{ll}
        \multicolumn{1}{l|}{task} & dataset \\
        \hline
        \multicolumn{1}{l|}{Short Caption} & \makecell[l]{Laion (en\&zh)~\cite{schuhmann2022laion}, COYO~\cite{kakaobrain2022coyo-700m}, COCO~\cite{lin2014microsoft}} \\
        
        \rowcolor{gray!15} \multicolumn{1}{l|}{OCR} & Wukong-OCR~\cite{gu2022wukong}, LaionCOCO-OCR~\cite{schuhmann2022laion-coco} \\
        \multicolumn{1}{l|}{Detection} & GRIT~\cite{peng2023kosmos} \\
        
        \rowcolor{gray!15} \multicolumn{1}{l|}{Conversation} & All-Seeing (en\&zh)~\cite{wang2023all} \\ 

        \multicolumn{2}{l}{Image-text instruction data (see Table~\ref{tab:ds_sft})} \\

    \end{tabular}
\vspace{-1em}
\caption{\textbf{Summary of datasets used in the alignment stage.}}
\label{tab:ds_align}

\label{tab:ds_pretrain}
\vspace{-1em}
\end{table}
\begin{table}[t]\scriptsize
\renewcommand{\arraystretch}{1.2}
\centering

\begin{tabular}{l|l}
         task & dataset \\
\hline
                                  & VQAv2~\cite{goyal2017making}, GQA~\cite{hudson2019gqa}, OKVQA~\cite{marino2019ok} \\
\multirow{-2}{*}{General QA}      & VSR~\cite{liu2023visual}  \\

\rowcolor{gray!15}
                                  & AI2D~\cite{kembhavi2016diagram}, ScienceQA~\cite{lu2022learn}, Chemistry Data~\cite{li2024chemvlm} \\
\rowcolor{gray!15}
\multirow{-2}{*}{Science}         & TQA~\cite{kembhavi2017you} \\

                                  & PMC-VQA~\cite{zhang2023pmc}, VQA-RAD~\cite{lau2018dataset}, VQA-Med~\cite{ben2019vqa} \\
                                  & Medical-Diff-VQA~\cite{hu2023medical}, PathVQA~\cite{he2020pathvqa}, 
 \\
\multirow{-3}{*}{Medical}         & SLAKE~\cite{liu2021slake}, PMC-CaseReport~\cite{pmc2023case} \\

\rowcolor{gray!15}
                                  & ChartQA~\cite{masry2022chartqa}, LRV-Instruction~\cite{liu2023mitigating}, PlotQA~\cite{methani2020plotqa} \\
\rowcolor{gray!15}
                                  & Unichart~\cite{masry2023unichart}, MMC-Inst~\cite{liu2023mmc}, DVQA~\cite{kafle2018dvqa} \\
\rowcolor{gray!15}
                                  & TableMWP~\cite{lu2022dynamic}, FigureQA~\cite{kahou2017figureqa}, MapQA~\cite{chang2022mapqa}  \\
\rowcolor{gray!15}
\multirow{-4}{*}{Chart}           & SciTSR~\cite{chi2019complicated}, Fintabnet~\cite{zheng2021global} \\

                                  &  CLEVR~\cite{johnson2017clevr}, MetaMath~\cite{yu2023metamath}, GeoQA+~\cite{cao2022augmented} \\
                                  & Geometry3k~\cite{lu2021inter}, GeoS~\cite{seo2015solving}, Unigeo~\cite{chen2022unigeo} \\
\multirow{-3}{*}{Mathematics}     & Super-CLEVR~\cite{li2023super}, MathQA~\cite{amini2019mathqa}\\

\rowcolor{gray!15}
                                  & Art500k~\cite{mao2017deepart}, MovieNet~\cite{huang2020movienet}, KonIQ-10k~\cite{hosu2020koniq} \\
\rowcolor{gray!15}
\multirow{-2}{*}{Knowledge}       & KVQA~\cite{shah2019kvqa}, ViQuAE~\cite{lerner2022viquae} \\

                                  & InfoVQA~\cite{mathew2022infographicvqa}, TextVQA~\cite{singh2019towards}, ArT~\cite{chng2019icdar2019} \\
                                  & CASIA~\cite{liu2011casia}, Chart-to-text~\cite{kantharaj2022chart}, COCO~\cite{veit2016coco} \\
                                  & CTW~\cite{yuan2019ctw}, EATEN~\cite{guo2019eaten}, ICDAR2019-LSVT~\cite{sun2019icdar} \\
                                  & ICPR MTWI~\cite{he2018icpr2018}, NAF~\cite{davis2019deep}, ReCTS~\cite{zhang2019icdar} \\
                                  & TextOCR~\cite{singh2021textocr}, LLaVAR~\cite{zhang2023llavar}, HME-100k~\cite{yuan2022syntax} \\
                                  & POIE~\cite{kuang2023visual}, SROIE~\cite{huang2019icdar2019}, ST-VQA~\cite{biten2019scene} \\
\multirow{-7}{*}{OCR}             & EST-VQA~\cite{wang2020general}, IAM~\cite{marti2002iam}\\

\rowcolor{gray!15}
Document                          & DocVQA~\cite{clark2017simple}, DocReason25k~\cite{hu2024mplug}\\

                                  & RefCOCO~\cite{kazemzadeh2014referitgame}, RefCOCO+~\cite{kazemzadeh2014referitgame}, RefCOCOg~\cite{kazemzadeh2014referitgame}  \\
\multirow{-2}{*}{Grounding}       & RD-BoxCoT~\cite{chen2023shikra} \\

\rowcolor{gray!15}
                                  & ALLaVA~\cite{chen2024allava}, LAION-GPT4V~\cite{laion_gpt4v_dataset} \\
\rowcolor{gray!15}
\multirow{-2}{*}{Conversation}    & MMDU~\cite{liu2024mmdu}, TextOCR-GPT4V~\cite{textocr-gpt4v} \\

Detection                         & Objects365~\cite{shao2019objects365}, V3Det~\cite{wang2023v3det} \\

\rowcolor{gray!15}
                                  & CLEVRER~\cite{yi2019clevrer}, EgoTaskQA~\cite{jia2022egotaskqa}, LSMDC~\cite{rohrbach2015dataset} \\
\rowcolor{gray!15}
                                  & Mementos~\cite{wang2024mementos}, STAR~\cite{wu2024star}, NTU RGB+D~\cite{shahroudy2016ntu} \\
\rowcolor{gray!15}
\multirow{-3}{*}{Video}           & VideoChat2-IT*~\cite{li2024mvbench}, VideoGPT+~\cite{maaz2024videogpt+}, \\
    \end{tabular}

\vspace{-1em}
\caption{\textbf{Summary of datasets used in the instruction tuning stage.} *IT refers to the instruction tuning data in VideoChat2.
}
\label{tab:ds_sft}
\end{table}

\section{More Ablation Studies}
\label{appendix:ablation}

\begin{figure*}
    \centering
    \includegraphics[width=\linewidth]{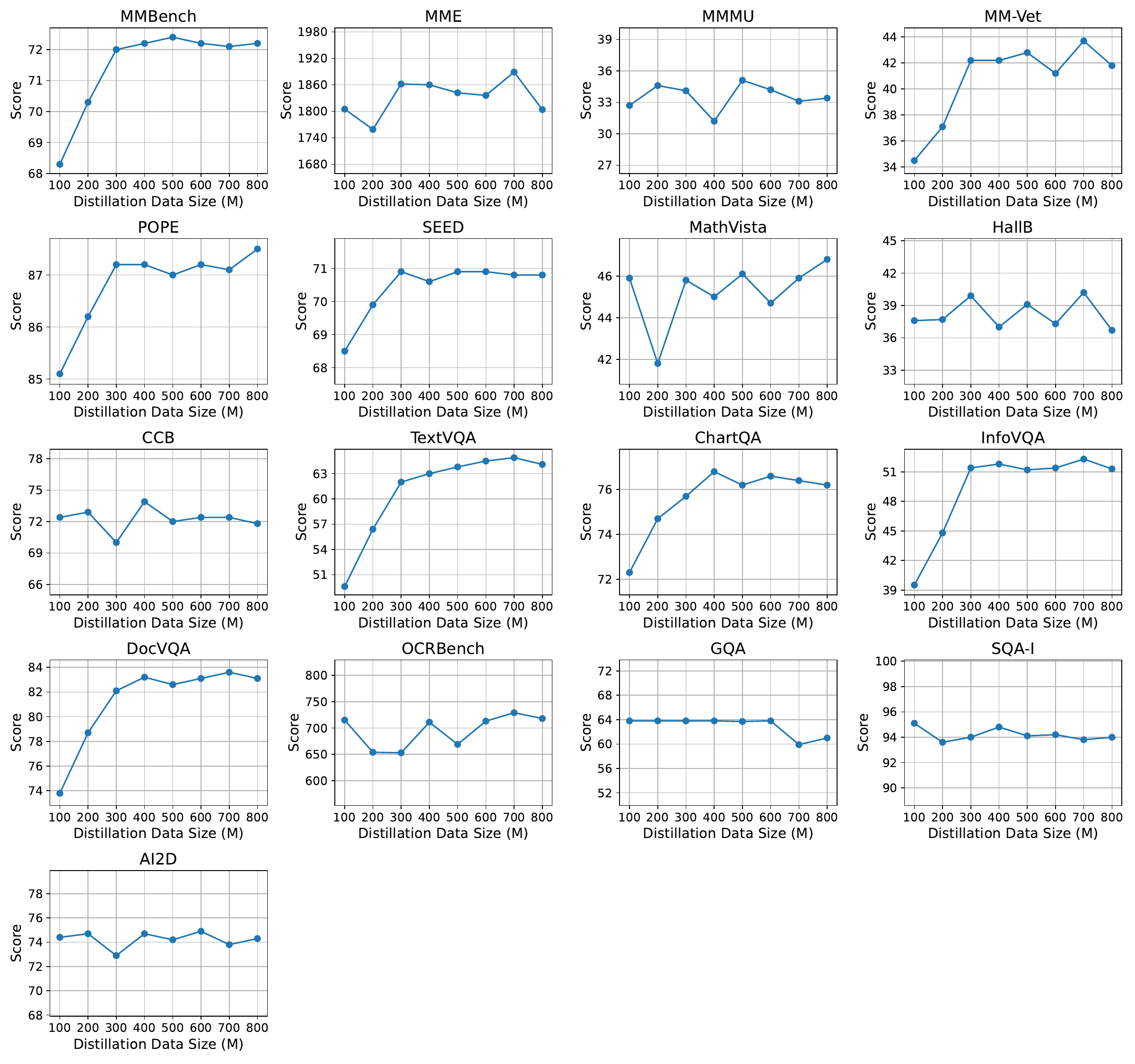}
    \vspace{-2em}
    \caption{\textbf{Distillation data scaling performance on 17 benchmarks.}}
    \label{fig:data_scaling}
\end{figure*}

\begin{table*}[t!]
\centering
\small
\setlength{\tabcolsep}{2.5pt}
\scalebox{0.95}{
\begin{tabular}{l c | c c c c c c c c c}
\toprule
Model & \makecell{Tile\\Resolution} &  MMB &  MME &  MMMU & MM-Vet & POPE & SEED & MathVista & HallB & CCB \\
\midrule
InternVL2 & $448\times 448$    & 73.2 & 1877 & 34.3 & 44.6 & 88.3 & 71.6 & 46.4 & 37.9 & 74.7 \\
\name{}~(HD) & $336\times 336$ & 72.9 & 1869 & 33.6 & 40.4 & 87.8 & 70.2 & 46.1 & 39.4 & 71.6 \\
\name{}~(HD) & $392\times 392$ & 72.3 & 1838 & 34.2 & 43.9 & 87.4 & 70.8 & 45.3 & 38.7 & 70.6 \\
\name{}~(HD) & $448\times 448$ & 73.3 & 1862 & 32.2 & 43.8 & 87.4 & 70.9 & 49.2 & 38.4 & 74.3 \\
\bottomrule
\end{tabular}
}
\vspace{-0.5em}
\caption{\textbf{Inference speed-performance trade-off of \name{}~(HD) on general VLM benchmarks.}}
\vspace{-1em}
\label{appendix:speed1}
\end{table*}

\begin{table*}[t!]
\centering
\small
\setlength{\tabcolsep}{2.5pt}
\scalebox{0.95}{
\begin{tabular}{l c | c c c c c c c c}
\toprule
Model & \makecell{Tile\\Resolution} & TextVQA &  ChartQA &  InfoVQA & DocVQA & OCRBench & GQA & SQA-I & AI2D \\
\midrule
InternVL2 & $448\times 448$    & 73.4 & 76.2 & 57.7 & 85.9 & 784 & 61.0 & 84.9 & 74.1 \\
\name{}~(HD) & $336\times 336$ & 68.1 & 76.7 & 55.6 & 84.2 & 739 & 60.8 & 94.6 & 73.1 \\
\name{}~(HD) & $392\times 392$ & 70.0 & 78.3 & 55.4 & 85.4 & 737 & 60.7 & 94.4 & 73.3 \\
\name{}~(HD) & $448\times 448$ & 70.9 & 78.6 & 55.7 & 86.1 & 740 & 60.4 & 94.8 & 73.0 \\
\bottomrule
\end{tabular}
}
\vspace{-0.5em}
\caption{\textbf{Inference speed-performance trade-off of \name{}~(HD) on visual question answering benchmarks.}}
\vspace{-2em}
\label{appendix:speed2}
\end{table*}

\noindent\textbf{Distillation Data Scaling.} We provide model performance change as data scales up on 17 benchmarks in Figure~\ref{fig:data_scaling}. It's shown that the model performance continues to improve on 9 benchmarks, while it oscillates on other 8 benchmarks. We hypothesize that the performance of these 8 benchmarks is bottle-necked by the LLM, not the holistic embedding. In contrast, the other benchmarks continue to benefit from additional distillation data, demonstrating the effectiveness of the distillation stage.

\vspace{0.5em}
\noindent\textbf{Speed-Performance Trade-off.} We provide the detailed performance of \name{}~(HD) with different tile resolutions in Table~\ref{appendix:speed1} and Table~\ref{appendix:speed2}. With tile resolution 336, \name{}~(HD) can already achieve comparable results with InternVL2. As the tile resolution increases, the performance of \name{}~(HD) steadily improves, especially on visual question answering benchmarks.

\begin{figure*}
    \centering
    \vspace{2em}
    \includegraphics[width=\linewidth]{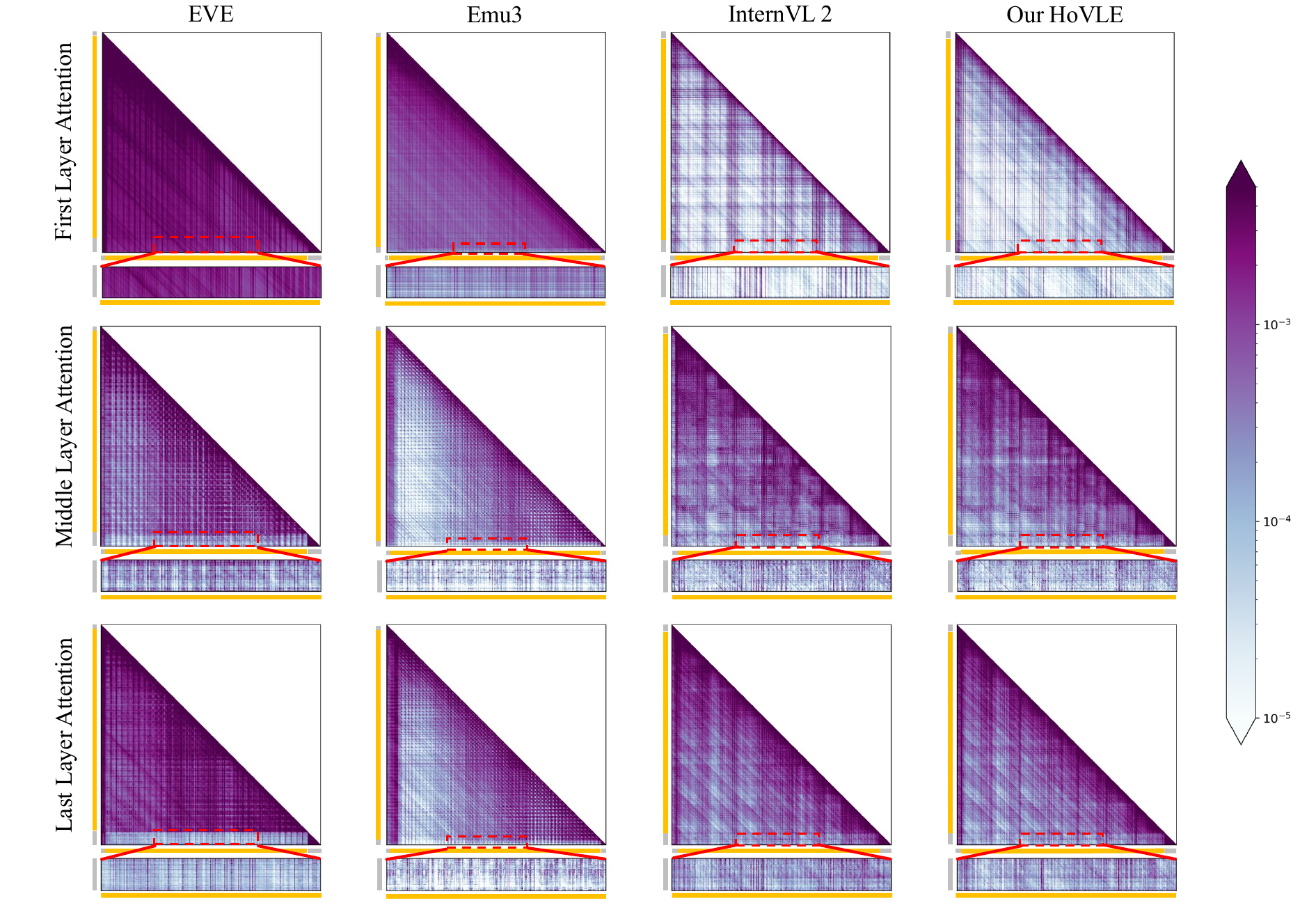}
    \caption{\textbf{Attention Maps for EVE, Emu3, InternVL2 and our HoVLE at the first, middle and last layers of LLM backbones}. Y-axis represents query tokens, and X-axis represents key tokens, with text modality tokens in gray and image modality tokens in yellow. All four models share the same input, but the sequence lengths of input tokens are different due to different image pre-processing. We highlight text-to-image attention below each full attention map. Our HoVLE, like the compositional InternVL2, has sparse attention across all network layers, while other monolithic models Emu3 and EVE have denser attention in shallow layers.}
    \label{fig:vis_mid}
    \vspace{-1em}
\end{figure*}

\section{Attention Map Visualization}
\label{appendix:vis}
Figure~\ref{fig:vis_mid} presents the visualization of attention map in the first, middle and last layers of EVE, Emu3, InternVL2 and our \name{}. It's shown that the text-to-image attention of previous monolithic models, like EVE and Emu3, displays dense pattern at the first layer, and gradually becomes sparse in deeper layers. On the contrary, compositional VLMs, like InternVL2, and our \name{} possess sparse text-to-image attention throughout all LLM layers.

\clearpage
{
    \small
    \bibliographystyle{ieeenat_fullname}
    \bibliography{main}

\begin{thebibliography}{135}
\providecommand{\natexlab}[1]{#1}
\providecommand{\url}[1]{\texttt{#1}}
\expandafter\ifx\csname urlstyle\endcsname\relax
  \providecommand{\doi}[1]{doi: #1}\else
  \providecommand{\doi}{doi: \begingroup \urlstyle{rm}\Url}\fi

\bibitem[Abdin et~al.(2024)Abdin, Aneja, Awadalla, Awadallah, Awan, Bach, Bahree, Bakhtiari, Bao, Behl, et~al.]{abdin2024phi}
Marah Abdin, Jyoti Aneja, Hany Awadalla, Ahmed Awadallah, Ammar~Ahmad Awan, Nguyen Bach, Amit Bahree, Arash Bakhtiari, Jianmin Bao, Harkirat Behl, et~al.
\newblock Phi-3 technical report: A highly capable language model locally on your phone.
\newblock \emph{arxiv:2404.14219}, 2024.

\bibitem[Achiam et~al.(2023)Achiam, Adler, Agarwal, Ahmad, Akkaya, Aleman, Almeida, Altenschmidt, Altman, Anadkat, et~al.]{achiam2023gpt}
Josh Achiam, Steven Adler, Sandhini Agarwal, Lama Ahmad, Ilge Akkaya, Florencia~Leoni Aleman, Diogo Almeida, Janko Altenschmidt, Sam Altman, Shyamal Anadkat, et~al.
\newblock Gpt-4 technical report.
\newblock \emph{arxiv:2303.08774}, 2023.

\bibitem[Amini et~al.(2019)Amini, Gabriel, Lin, Koncel-Kedziorski, Choi, and Hajishirzi]{amini2019mathqa}
Aida Amini, Saadia Gabriel, Peter Lin, Rik Koncel-Kedziorski, Yejin Choi, and Hannaneh Hajishirzi.
\newblock Mathqa: Towards interpretable math word problem solving with operation-based formalisms.
\newblock \emph{arXiv preprint arXiv:1905.13319}, 2019.

\bibitem[Anthropic(2024)]{TheC3}
AI Anthropic.
\newblock The claude 3 model family: Opus, sonnet, haiku.
\newblock Claude-3 Model Card, 2024.

\bibitem[Bai et~al.(2023)Bai, Bai, Yang, Wang, Tan, Wang, Lin, Zhou, and Zhou]{bai2023qwen}
Jinze Bai, Shuai Bai, Shusheng Yang, Shijie Wang, Sinan Tan, Peng Wang, Junyang Lin, Chang Zhou, and Jingren Zhou.
\newblock Qwen-vl: A frontier large vision-language model with versatile abilities.
\newblock \emph{arxiv:2308.12966}, 2023.

\bibitem[Bavishi et~al.(2023)Bavishi, Elsen, Hawthorne, Nye, Odena, Somani, and Ta\c{s}\i{}rlar]{fuyu-8b}
Rohan Bavishi, Erich Elsen, Curtis Hawthorne, Maxwell Nye, Augustus Odena, Arushi Somani, and Sa\u{g}nak Ta\c{s}\i{}rlar.
\newblock Introducing our multimodal models.
\newblock \url{https://www.adept.ai/blog/fuyu-8b}, 2023.

\bibitem[Ben~Abacha et~al.(2019)Ben~Abacha, Hasan, Datla, Demner-Fushman, and M{\"u}ller]{ben2019vqa}
Asma Ben~Abacha, Sadid~A Hasan, Vivek~V Datla, Dina Demner-Fushman, and Henning M{\"u}ller.
\newblock Vqa-med: Overview of the medical visual question answering task at imageclef 2019.
\newblock In \emph{Proceedings of CLEF (Conference and Labs of the Evaluation Forum) 2019 Working Notes}. 9-12 September 2019, 2019.

\bibitem[Beyer et~al.(2024)Beyer, Steiner, Pinto, Kolesnikov, Wang, Salz, Neumann, Alabdulmohsin, Tschannen, Bugliarello, et~al.]{beyer2024paligemma}
Lucas Beyer, Andreas Steiner, Andr{\'e}~Susano Pinto, Alexander Kolesnikov, Xiao Wang, Daniel Salz, Maxim Neumann, Ibrahim Alabdulmohsin, Michael Tschannen, Emanuele Bugliarello, et~al.
\newblock Paligemma: A versatile 3b vlm for transfer.
\newblock \emph{arxiv:2407.07726}, 2024.

\bibitem[Biten et~al.(2019)Biten, Tito, Mafla, Gomez, Rusinol, Valveny, Jawahar, and Karatzas]{biten2019scene}
Ali~Furkan Biten, Ruben Tito, Andres Mafla, Lluis Gomez, Mar{\c{c}}al Rusinol, Ernest Valveny, CV Jawahar, and Dimosthenis Karatzas.
\newblock Scene text visual question answering.
\newblock In \emph{Proceedings of the IEEE/CVF international conference on computer vision}, pages 4291--4301, 2019.

\bibitem[Byeon et~al.(2022)Byeon, Park, Kim, Lee, Baek, and Kim]{kakaobrain2022coyo-700m}
Minwoo Byeon, Beomhee Park, Haecheon Kim, Sungjun Lee, Woonhyuk Baek, and Saehoon Kim.
\newblock Coyo-700m: Image-text pair dataset.
\newblock \url{https://github.com/kakaobrain/coyo-dataset}, 2022.

\bibitem[Cai et~al.(2024)Cai, Cao, Chen, Chen, Chen, Chen, Chen, Chen, Chen, Chu, et~al.]{cai2024internlm2}
Zheng Cai, Maosong Cao, Haojiong Chen, Kai Chen, Keyu Chen, Xin Chen, Xun Chen, Zehui Chen, Zhi Chen, Pei Chu, et~al.
\newblock Internlm2 technical report.
\newblock \emph{arxiv:2403.17297}, 2024.

\bibitem[Cao and Xiao(2022)]{cao2022augmented}
Jie Cao and Jing Xiao.
\newblock An augmented benchmark dataset for geometric question answering through dual parallel text encoding.
\newblock In \emph{Proceedings of the 29th International Conference on Computational Linguistics}, pages 1511--1520, 2022.

\bibitem[Carter(2024)]{textocr-gpt4v}
Jimmy Carter.
\newblock Textocr-gpt4v.
\newblock \url{https://huggingface.co/datasets/jimmycarter/textocr-gpt4v}, 2024.

\bibitem[Chang et~al.(2022)Chang, Palzer, Li, Fosler-Lussier, and Xiao]{chang2022mapqa}
Shuaichen Chang, David Palzer, Jialin Li, Eric Fosler-Lussier, and Ningchuan Xiao.
\newblock Mapqa: A dataset for question answering on choropleth maps.
\newblock \emph{arXiv preprint arXiv:2211.08545}, 2022.

\bibitem[Chen et~al.(2024{\natexlab{a}})Chen, Chen, Zhang, Chen, Wu, Zhang, Chen, Li, Wan, and Wang]{chen2024allava}
Guiming~Hardy Chen, Shunian Chen, Ruifei Zhang, Junying Chen, Xiangbo Wu, Zhiyi Zhang, Zhihong Chen, Jianquan Li, Xiang Wan, and Benyou Wang.
\newblock Allava: Harnessing gpt4v-synthesized data for a lite vision-language model.
\newblock \emph{arXiv preprint arXiv:2402.11684}, 2024{\natexlab{a}}.

\bibitem[Chen et~al.(2022)Chen, Li, Qin, Lu, Lin, Chen, and Liang]{chen2022unigeo}
Jiaqi Chen, Tong Li, Jinghui Qin, Pan Lu, Liang Lin, Chongyu Chen, and Xiaodan Liang.
\newblock Unigeo: Unifying geometry logical reasoning via reformulating mathematical expression.
\newblock \emph{arXiv preprint arXiv:2212.02746}, 2022.

\bibitem[Chen et~al.(2023)Chen, Zhang, Zeng, Zhang, Zhu, and Zhao]{chen2023shikra}
Keqin Chen, Zhao Zhang, Weili Zeng, Richong Zhang, Feng Zhu, and Rui Zhao.
\newblock Shikra: Unleashing multimodal llm's referential dialogue magic.
\newblock \emph{arXiv preprint arXiv:2306.15195}, 2023.

\bibitem[Chen et~al.(2024{\natexlab{b}})Chen, Wang, Peng, and Ji]{chen2024single}
Yangyi Chen, Xingyao Wang, Hao Peng, and Heng Ji.
\newblock A single transformer for scalable vision-language modeling.
\newblock \emph{arxiv:2407.06438}, 2024{\natexlab{b}}.

\bibitem[Chen et~al.(2024{\natexlab{c}})Chen, Wang, Tian, Ye, Gao, Cui, Tong, Hu, Luo, Ma, et~al.]{chen2024far}
Zhe Chen, Weiyun Wang, Hao Tian, Shenglong Ye, Zhangwei Gao, Erfei Cui, Wenwen Tong, Kongzhi Hu, Jiapeng Luo, Zheng Ma, et~al.
\newblock How far are we to gpt-4v? closing the gap to commercial multimodal models with open-source suites.
\newblock \emph{arxiv:2404.16821}, 2024{\natexlab{c}}.

\bibitem[Chen et~al.(2024{\natexlab{d}})Chen, Wang, Tian, Ye, Gao, Cui, Tong, Hu, Luo, Ma, et~al.]{internvl2}
Zhe Chen, Weiyun Wang, Hao Tian, Shenglong Ye, Zhangwei Gao, Erfei Cui, Wenwen Tong, Kongzhi Hu, Jiapeng Luo, Zheng Ma, et~al.
\newblock Internvl2: Better than the best—expanding performance boundaries of open-source multimodal models with the progressive scaling strategy.
\newblock \url{https://internvl.github.io/blog/2024-07-02-InternVL-2.0/}, 2024{\natexlab{d}}.

\bibitem[Chen et~al.(2024{\natexlab{e}})Chen, Wu, Wang, Su, Chen, Xing, Zhong, Zhang, Zhu, Lu, et~al.]{chen2024internvl}
Zhe Chen, Jiannan Wu, Wenhai Wang, Weijie Su, Guo Chen, Sen Xing, Muyan Zhong, Qinglong Zhang, Xizhou Zhu, Lewei Lu, et~al.
\newblock Internvl: Scaling up vision foundation models and aligning for generic visual-linguistic tasks.
\newblock In \emph{CVPR}, pages 24185--24198, 2024{\natexlab{e}}.

\bibitem[Chi et~al.(2019)Chi, Huang, Xu, Yu, Yin, and Mao]{chi2019complicated}
Zewen Chi, Heyan Huang, Heng-Da Xu, Houjin Yu, Wanxuan Yin, and Xian-Ling Mao.
\newblock Complicated table structure recognition.
\newblock \emph{arXiv preprint arXiv:1908.04729}, 2019.

\bibitem[Chng et~al.(2019)Chng, Liu, Sun, Ng, Luo, Ni, Fang, Zhang, Han, Ding, et~al.]{chng2019icdar2019}
Chee~Kheng Chng, Yuliang Liu, Yipeng Sun, Chun~Chet Ng, Canjie Luo, Zihan Ni, ChuanMing Fang, Shuaitao Zhang, Junyu Han, Errui Ding, et~al.
\newblock Icdar2019 robust reading challenge on arbitrary-shaped text-rrc-art.
\newblock In \emph{2019 International Conference on Document Analysis and Recognition (ICDAR)}, pages 1571--1576. IEEE, 2019.

\bibitem[Clark and Gardner(2017)]{clark2017simple}
Christopher Clark and Matt Gardner.
\newblock Simple and effective multi-paragraph reading comprehension.
\newblock \emph{arXiv preprint arXiv:1710.10723}, 2017.

\bibitem[Davis et~al.(2019)Davis, Morse, Cohen, Price, and Tensmeyer]{davis2019deep}
Brian Davis, Bryan Morse, Scott Cohen, Brian Price, and Chris Tensmeyer.
\newblock Deep visual template-free form parsing.
\newblock In \emph{2019 International Conference on Document Analysis and Recognition (ICDAR)}, pages 134--141. IEEE, 2019.

\bibitem[Diao et~al.(2024)Diao, Cui, Li, Wang, Lu, and Wang]{diao2024unveiling}
Haiwen Diao, Yufeng Cui, Xiaotong Li, Yueze Wang, Huchuan Lu, and Xinlong Wang.
\newblock Unveiling encoder-free vision-language models.
\newblock \emph{arxiv:2406.11832}, 2024.

\bibitem[Fang et~al.(2021)Fang, Wang, Hu, Wang, Yang, and Liu]{fang2021compressing}
Zhiyuan Fang, Jianfeng Wang, Xiaowei Hu, Lijuan Wang, Yezhou Yang, and Zicheng Liu.
\newblock Compressing visual-linguistic model via knowledge distillation.
\newblock In \emph{ICCV}, pages 1428--1438, 2021.

\bibitem[Fu et~al.(2023)Fu, Chen, Shen, Qin, Zhang, Lin, Yang, Zheng, Li, Sun, et~al.]{fu2024mme}
Chaoyou Fu, Peixian Chen, Yunhang Shen, Yulei Qin, Mengdan Zhang, Xu Lin, Jinrui Yang, Xiawu Zheng, Ke Li, Xing Sun, et~al.
\newblock Mme: A comprehensive evaluation benchmark for multimodal large language models.
\newblock \emph{arxiv:2306.13394}, 2023.

\bibitem[Goyal et~al.(2017)Goyal, Khot, Summers-Stay, Batra, and Parikh]{goyal2017making}
Yash Goyal, Tejas Khot, Douglas Summers-Stay, Dhruv Batra, and Devi Parikh.
\newblock Making the v in vqa matter: Elevating the role of image understanding in visual question answering.
\newblock In \emph{Proceedings of the IEEE conference on computer vision and pattern recognition}, pages 6904--6913, 2017.

\bibitem[Gu et~al.(2022)Gu, Meng, Lu, Hou, Minzhe, Liang, Yao, Huang, Zhang, Jiang, et~al.]{gu2022wukong}
Jiaxi Gu, Xiaojun Meng, Guansong Lu, Lu Hou, Niu Minzhe, Xiaodan Liang, Lewei Yao, Runhui Huang, Wei Zhang, Xin Jiang, et~al.
\newblock Wukong: A 100 million large-scale chinese cross-modal pre-training benchmark.
\newblock \emph{Advances in Neural Information Processing Systems}, 35:\penalty0 26418--26431, 2022.

\bibitem[Guan et~al.(2023)Guan, Liu, Wu, Xian, Li, Liu, Wang, Chen, Huang, Yacoob, et~al.]{guan2023hallusionbench}
Tianrui Guan, Fuxiao Liu, Xiyang Wu, Ruiqi Xian, Zongxia Li, Xiaoyu Liu, Xijun Wang, Lichang Chen, Furong Huang, Yaser Yacoob, et~al.
\newblock Hallusionbench: An advanced diagnostic suite for entangled language hallucination and visual illusion in large vision-language models.
\newblock \emph{arxiv:2310.14566}, 2023.

\bibitem[Guo et~al.(2019)Guo, Qin, Liu, Han, Liu, and Ding]{guo2019eaten}
He Guo, Xiameng Qin, Jiaming Liu, Junyu Han, Jingtuo Liu, and Errui Ding.
\newblock Eaten: Entity-aware attention for single shot visual text extraction.
\newblock In \emph{2019 International Conference on Document Analysis and Recognition (ICDAR)}, pages 254--259. IEEE, 2019.

\bibitem[Gururangan et~al.(2020)Gururangan, Marasovi{\'c}, Swayamdipta, Lo, Beltagy, Downey, and Smith]{gururangan2020don}
Suchin Gururangan, Ana Marasovi{\'c}, Swabha Swayamdipta, Kyle Lo, Iz Beltagy, Doug Downey, and Noah~A Smith.
\newblock Don’t stop pretraining: Adapt language models to domains and tasks.
\newblock In \emph{ACL}, pages 8342--8360, 2020.

\bibitem[He et~al.(2018)He, Liu, Yang, Zhang, Luo, Gao, Zheng, Wang, Zhang, and Jin]{he2018icpr2018}
Mengchao He, Yuliang Liu, Zhibo Yang, Sheng Zhang, Canjie Luo, Feiyu Gao, Qi Zheng, Yongpan Wang, Xin Zhang, and Lianwen Jin.
\newblock Icpr2018 contest on robust reading for multi-type web images.
\newblock In \emph{2018 24th international conference on pattern recognition (ICPR)}, pages 7--12. IEEE, 2018.

\bibitem[He et~al.(2020)He, Zhang, Mou, Xing, and Xie]{he2020pathvqa}
Xuehai He, Yichen Zhang, Luntian Mou, Eric Xing, and Pengtao Xie.
\newblock Pathvqa: 30000+ questions for medical visual question answering.
\newblock \emph{arXiv preprint arXiv:2003.10286}, 2020.

\bibitem[Hosu et~al.(2020)Hosu, Lin, Sziranyi, and Saupe]{hosu2020koniq}
Vlad Hosu, Hanhe Lin, Tamas Sziranyi, and Dietmar Saupe.
\newblock Koniq-10k: An ecologically valid database for deep learning of blind image quality assessment.
\newblock \emph{IEEE Transactions on Image Processing}, 29:\penalty0 4041--4056, 2020.

\bibitem[Hu et~al.(2024{\natexlab{a}})Hu, Xu, Ye, Yan, Zhang, Zhang, Li, Zhang, Jin, Huang, et~al.]{hu2024mplug}
Anwen Hu, Haiyang Xu, Jiabo Ye, Ming Yan, Liang Zhang, Bo Zhang, Chen Li, Ji Zhang, Qin Jin, Fei Huang, et~al.
\newblock mplug-docowl 1.5: Unified structure learning for ocr-free document understanding.
\newblock \emph{arXiv preprint arXiv:2403.12895}, 2024{\natexlab{a}}.

\bibitem[Hu et~al.(2023)Hu, Gu, An, Zhang, Liu, Kobayashi, Harada, Summers, and Zhu]{hu2023medical}
Xinyue Hu, L Gu, Q An, M Zhang, L Liu, K Kobayashi, T Harada, R Summers, and Y Zhu.
\newblock Medical-diff-vqa: a large-scale medical dataset for difference visual question answering on chest x-ray images, 2023.

\bibitem[Hu et~al.(2024{\natexlab{b}})Hu, Stretcu, Lu, Viswanathan, Hata, Luo, Krishna, and Fuxman]{hu2024visual}
Yushi Hu, Otilia Stretcu, Chun-Ta Lu, Krishnamurthy Viswanathan, Kenji Hata, Enming Luo, Ranjay Krishna, and Ariel Fuxman.
\newblock Visual program distillation: Distilling tools and programmatic reasoning into vision-language models.
\newblock In \emph{CVPR}, pages 9590--9601, 2024{\natexlab{b}}.

\bibitem[Huang et~al.(2020)Huang, Xiong, Rao, Wang, and Lin]{huang2020movienet}
Qingqiu Huang, Yu Xiong, Anyi Rao, Jiaze Wang, and Dahua Lin.
\newblock Movienet: A holistic dataset for movie understanding.
\newblock In \emph{Computer Vision--ECCV 2020: 16th European Conference, Glasgow, UK, August 23--28, 2020, Proceedings, Part IV 16}, pages 709--727. Springer, 2020.

\bibitem[Huang et~al.(2024)Huang, Ding, Wang, Han, Liu, Zhao, Xu, Hou, Zhang, and Liang]{huang2024hires}
Runhui Huang, Xinpeng Ding, Chunwei Wang, Jianhua Han, Yulong Liu, Hengshuang Zhao, Hang Xu, Lu Hou, Wei Zhang, and Xiaodan Liang.
\newblock Hires-llava: Restoring fragmentation input in high-resolution large vision-language models.
\newblock \emph{arXiv:2407.08706}, 2024.

\bibitem[Huang et~al.(2019)Huang, Chen, He, Bai, Karatzas, Lu, and Jawahar]{huang2019icdar2019}
Zheng Huang, Kai Chen, Jianhua He, Xiang Bai, Dimosthenis Karatzas, Shijian Lu, and CV Jawahar.
\newblock Icdar2019 competition on scanned receipt ocr and information extraction.
\newblock In \emph{2019 International Conference on Document Analysis and Recognition (ICDAR)}, pages 1516--1520. IEEE, 2019.

\bibitem[Hudson and Manning(2019)]{hudson2019gqa}
Drew~A Hudson and Christopher~D Manning.
\newblock Gqa: A new dataset for real-world visual reasoning and compositional question answering.
\newblock In \emph{Proceedings of the IEEE/CVF conference on computer vision and pattern recognition}, pages 6700--6709, 2019.

\bibitem[Jia et~al.(2022)Jia, Lei, Zhu, and Huang]{jia2022egotaskqa}
Baoxiong Jia, Ting Lei, Song-Chun Zhu, and Siyuan Huang.
\newblock Egotaskqa: Understanding human tasks in egocentric videos.
\newblock \emph{Advances in Neural Information Processing Systems}, 35:\penalty0 3343--3360, 2022.

\bibitem[Johnson et~al.(2017)Johnson, Hariharan, Van Der~Maaten, Fei-Fei, Lawrence~Zitnick, and Girshick]{johnson2017clevr}
Justin Johnson, Bharath Hariharan, Laurens Van Der~Maaten, Li Fei-Fei, C Lawrence~Zitnick, and Ross Girshick.
\newblock Clevr: A diagnostic dataset for compositional language and elementary visual reasoning.
\newblock In \emph{Proceedings of the IEEE conference on computer vision and pattern recognition}, pages 2901--2910, 2017.

\bibitem[Kafle et~al.(2018)Kafle, Price, Cohen, and Kanan]{kafle2018dvqa}
Kushal Kafle, Brian Price, Scott Cohen, and Christopher Kanan.
\newblock Dvqa: Understanding data visualizations via question answering.
\newblock In \emph{Proceedings of the IEEE conference on computer vision and pattern recognition}, pages 5648--5656, 2018.

\bibitem[Kahou et~al.(2017)Kahou, Michalski, Atkinson, K{\'a}d{\'a}r, Trischler, and Bengio]{kahou2017figureqa}
Samira~Ebrahimi Kahou, Vincent Michalski, Adam Atkinson, {\'A}kos K{\'a}d{\'a}r, Adam Trischler, and Yoshua Bengio.
\newblock Figureqa: An annotated figure dataset for visual reasoning.
\newblock \emph{arXiv preprint arXiv:1710.07300}, 2017.

\bibitem[Kantharaj et~al.(2022)Kantharaj, Leong, Lin, Masry, Thakkar, Hoque, and Joty]{kantharaj2022chart}
Shankar Kantharaj, Rixie Tiffany~Ko Leong, Xiang Lin, Ahmed Masry, Megh Thakkar, Enamul Hoque, and Shafiq Joty.
\newblock Chart-to-text: A large-scale benchmark for chart summarization.
\newblock \emph{arXiv preprint arXiv:2203.06486}, 2022.

\bibitem[Kazemzadeh et~al.(2014)Kazemzadeh, Ordonez, Matten, and Berg]{kazemzadeh2014referitgame}
Sahar Kazemzadeh, Vicente Ordonez, Mark Matten, and Tamara Berg.
\newblock Referitgame: Referring to objects in photographs of natural scenes.
\newblock In \emph{Proceedings of the 2014 conference on empirical methods in natural language processing (EMNLP)}, pages 787--798, 2014.

\bibitem[Kembhavi et~al.(2016)Kembhavi, Salvato, Kolve, Seo, Hajishirzi, and Farhadi]{kembhavi2016diagram}
Aniruddha Kembhavi, Mike Salvato, Eric Kolve, Minjoon Seo, Hannaneh Hajishirzi, and Ali Farhadi.
\newblock A diagram is worth a dozen images.
\newblock In \emph{ECCV}, pages 235--251, 2016.

\bibitem[Kembhavi et~al.(2017)Kembhavi, Seo, Schwenk, Choi, Farhadi, and Hajishirzi]{kembhavi2017you}
Aniruddha Kembhavi, Minjoon Seo, Dustin Schwenk, Jonghyun Choi, Ali Farhadi, and Hannaneh Hajishirzi.
\newblock Are you smarter than a sixth grader? textbook question answering for multimodal machine comprehension.
\newblock In \emph{Proceedings of the IEEE Conference on Computer Vision and Pattern recognition}, pages 4999--5007, 2017.

\bibitem[Kuang et~al.(2023)Kuang, Hua, Liang, Yang, Jiang, Ren, and Bai]{kuang2023visual}
Jianfeng Kuang, Wei Hua, Dingkang Liang, Mingkun Yang, Deqiang Jiang, Bo Ren, and Xiang Bai.
\newblock Visual information extraction in the wild: practical dataset and end-to-end solution.
\newblock In \emph{International Conference on Document Analysis and Recognition}, pages 36--53. Springer, 2023.

\bibitem[LAION(2023)]{laion_gpt4v_dataset}
LAION.
\newblock Laion-gpt4v dataset.
\newblock \url{https://huggingface.co/datasets/laion/gpt4v-dataset}, 2023.

\bibitem[Lau et~al.(2018)Lau, Gayen, Ben~Abacha, and Demner-Fushman]{lau2018dataset}
Jason~J Lau, Soumya Gayen, Asma Ben~Abacha, and Dina Demner-Fushman.
\newblock A dataset of clinically generated visual questions and answers about radiology images.
\newblock \emph{Scientific data}, 5\penalty0 (1):\penalty0 1--10, 2018.

\bibitem[Lerner et~al.(2022)Lerner, Ferret, Guinaudeau, Le~Borgne, Besan{\c{c}}on, Moreno, and Lov{\'o}n~Melgarejo]{lerner2022viquae}
Paul Lerner, Olivier Ferret, Camille Guinaudeau, Herv{\'e} Le~Borgne, Romaric Besan{\c{c}}on, Jos{\'e}~G Moreno, and Jes{\'u}s Lov{\'o}n~Melgarejo.
\newblock Viquae, a dataset for knowledge-based visual question answering about named entities.
\newblock In \emph{Proceedings of the 45th International ACM SIGIR Conference on Research and Development in Information Retrieval}, pages 3108--3120, 2022.

\bibitem[Li et~al.(2023{\natexlab{a}})Li, Wang, Wang, Ge, Ge, and Shan]{li2023seed}
Bohao Li, Rui Wang, Guangzhi Wang, Yuying Ge, Yixiao Ge, and Ying Shan.
\newblock Seed-bench: Benchmarking multimodal llms with generative comprehension.
\newblock \emph{arxiv:2307.16125}, 2023{\natexlab{a}}.

\bibitem[Li et~al.(2022)Li, Li, Xiong, and Hoi]{li2022blip}
Junnan Li, Dongxu Li, Caiming Xiong, and Steven Hoi.
\newblock Blip: Bootstrapping language-image pre-training for unified vision-language understanding and generation.
\newblock In \emph{ICML}, pages 12888--12900. PMLR, 2022.

\bibitem[Li et~al.(2024{\natexlab{a}})Li, Zhang, Wang, Hao, Lei, Tan, Zhou, Liu, Yang, Xiong, et~al.]{li2024chemvlm}
Junxian Li, Di Zhang, Xunzhi Wang, Zeying Hao, Jingdi Lei, Qian Tan, Cai Zhou, Wei Liu, Yaotian Yang, Xinrui Xiong, et~al.
\newblock Chemvlm: Exploring the power of multimodal large language models in chemistry area.
\newblock \emph{arXiv preprint arXiv:2408.07246}, 2024{\natexlab{a}}.

\bibitem[Li et~al.(2024{\natexlab{b}})Li, Wang, He, Li, Wang, Liu, Wang, Xu, Chen, Luo, et~al.]{li2024mvbench}
Kunchang Li, Yali Wang, Yinan He, Yizhuo Li, Yi Wang, Yi Liu, Zun Wang, Jilan Xu, Guo Chen, Ping Luo, et~al.
\newblock Mvbench: A comprehensive multi-modal video understanding benchmark.
\newblock In \emph{CVPR}, pages 22195--22206, 2024{\natexlab{b}}.

\bibitem[Li et~al.(2023{\natexlab{b}})Li, Du, Zhou, Wang, Zhao, and Wen]{li2023evaluating}
Yifan Li, Yifan Du, Kun Zhou, Jinpeng Wang, Wayne~Xin Zhao, and Ji-Rong Wen.
\newblock Evaluating object hallucination in large vision-language models.
\newblock In \emph{EMNLP}, pages 292--305, 2023{\natexlab{b}}.

\bibitem[Li et~al.(2023{\natexlab{c}})Li, Wang, Stengel-Eskin, Kortylewski, Ma, Van~Durme, and Yuille]{li2023super}
Zhuowan Li, Xingrui Wang, Elias Stengel-Eskin, Adam Kortylewski, Wufei Ma, Benjamin Van~Durme, and Alan~L Yuille.
\newblock Super-clevr: A virtual benchmark to diagnose domain robustness in visual reasoning.
\newblock In \emph{Proceedings of the IEEE/CVF Conference on Computer Vision and Pattern Recognition}, pages 14963--14973, 2023{\natexlab{c}}.

\bibitem[Li et~al.(2024{\natexlab{c}})Li, Yang, Liu, Ma, Zhang, Yang, Sun, Liu, and Bai]{li2024monkey}
Zhang Li, Biao Yang, Qiang Liu, Zhiyin Ma, Shuo Zhang, Jingxu Yang, Yabo Sun, Yuliang Liu, and Xiang Bai.
\newblock Monkey: Image resolution and text label are important things for large multi-modal models.
\newblock In \emph{CVPR}, pages 26763--26773, 2024{\natexlab{c}}.

\bibitem[Lin et~al.(2024)Lin, Yin, Ping, Molchanov, Shoeybi, and Han]{lin2024vila}
Ji Lin, Hongxu Yin, Wei Ping, Pavlo Molchanov, Mohammad Shoeybi, and Song Han.
\newblock Vila: On pre-training for visual language models.
\newblock In \emph{CVPR}, pages 26689--26699, 2024.

\bibitem[Lin et~al.(2014)Lin, Maire, Belongie, Hays, Perona, Ramanan, Doll{\'a}r, and Zitnick]{lin2014microsoft}
Tsung-Yi Lin, Michael Maire, Serge Belongie, James Hays, Pietro Perona, Deva Ramanan, Piotr Doll{\'a}r, and C~Lawrence Zitnick.
\newblock Microsoft coco: Common objects in context.
\newblock In \emph{Computer Vision--ECCV 2014: 13th European Conference, Zurich, Switzerland, September 6-12, 2014, Proceedings, Part V 13}, pages 740--755. Springer, 2014.

\bibitem[Liu et~al.(2021)Liu, Zhan, Xu, Ma, Yang, and Wu]{liu2021slake}
Bo Liu, Li-Ming Zhan, Li Xu, Lin Ma, Yan Yang, and Xiao-Ming Wu.
\newblock Slake: A semantically-labeled knowledge-enhanced dataset for medical visual question answering.
\newblock In \emph{2021 IEEE 18th International Symposium on Biomedical Imaging (ISBI)}, pages 1650--1654. IEEE, 2021.

\bibitem[Liu et~al.(2011)Liu, Yin, Wang, and Wang]{liu2011casia}
Cheng-Lin Liu, Fei Yin, Da-Han Wang, and Qiu-Feng Wang.
\newblock Casia online and offline chinese handwriting databases.
\newblock In \emph{2011 international conference on document analysis and recognition}, pages 37--41. IEEE, 2011.

\bibitem[Liu et~al.(2023{\natexlab{a}})Liu, Emerson, and Collier]{liu2023visual}
Fangyu Liu, Guy Emerson, and Nigel Collier.
\newblock Visual spatial reasoning.
\newblock \emph{Transactions of the Association for Computational Linguistics}, 11:\penalty0 635--651, 2023{\natexlab{a}}.

\bibitem[Liu et~al.(2023{\natexlab{b}})Liu, Lin, Li, Wang, Yacoob, and Wang]{liu2023mitigating}
Fuxiao Liu, Kevin Lin, Linjie Li, Jianfeng Wang, Yaser Yacoob, and Lijuan Wang.
\newblock Mitigating hallucination in large multi-modal models via robust instruction tuning.
\newblock In \emph{The Twelfth International Conference on Learning Representations}, 2023{\natexlab{b}}.

\bibitem[Liu et~al.(2023{\natexlab{c}})Liu, Wang, Yao, Chen, Song, Cho, Yacoob, and Yu]{liu2023mmc}
Fuxiao Liu, Xiaoyang Wang, Wenlin Yao, Jianshu Chen, Kaiqiang Song, Sangwoo Cho, Yaser Yacoob, and Dong Yu.
\newblock Mmc: Advancing multimodal chart understanding with large-scale instruction tuning.
\newblock \emph{arXiv preprint arXiv:2311.10774}, 2023{\natexlab{c}}.

\bibitem[Liu et~al.(2024{\natexlab{a}})Liu, Li, Li, and Lee]{liu2024improved}
Haotian Liu, Chunyuan Li, Yuheng Li, and Yong~Jae Lee.
\newblock Improved baselines with visual instruction tuning.
\newblock In \emph{CVPR}, pages 26296--26306, 2024{\natexlab{a}}.

\bibitem[Liu et~al.(2024{\natexlab{b}})Liu, Li, Li, Li, Zhang, Shen, and Lee]{liu2024llavanext}
Haotian Liu, Chunyuan Li, Yuheng Li, Bo Li, Yuanhan Zhang, Sheng Shen, and Yong~Jae Lee.
\newblock Llava-next: Improved reasoning, ocr, and world knowledge.
\newblock \url{https://llava-vl.github.io/blog/2024-01-30-llava-next/}, 2024{\natexlab{b}}.

\bibitem[Liu et~al.(2024{\natexlab{c}})Liu, Li, Wu, and Lee]{liu2024visual}
Haotian Liu, Chunyuan Li, Qingyang Wu, and Yong~Jae Lee.
\newblock Visual instruction tuning.
\newblock In \emph{NeurIPS}, 2024{\natexlab{c}}.

\bibitem[Liu et~al.(2023{\natexlab{d}})Liu, Li, Yang, Li, Yin, Liu, Jin, and Bai]{liu2023hidden}
Yuliang Liu, Zhang Li, Biao Yang, Chunyuan Li, Xucheng Yin, Cheng-lin Liu, Lianwen Jin, and Xiang Bai.
\newblock On the hidden mystery of ocr in large multimodal models.
\newblock \emph{arxiv:2305.07895}, 2023{\natexlab{d}}.

\bibitem[Liu et~al.(2025)Liu, Duan, Zhang, Li, Zhang, Zhao, Yuan, Wang, He, Liu, et~al.]{liu2025mmbench}
Yuan Liu, Haodong Duan, Yuanhan Zhang, Bo Li, Songyang Zhang, Wangbo Zhao, Yike Yuan, Jiaqi Wang, Conghui He, Ziwei Liu, et~al.
\newblock Mmbench: Is your multi-modal model an all-around player?
\newblock In \emph{ECCV}, pages 216--233. Springer, 2025.

\bibitem[Liu et~al.(2024{\natexlab{d}})Liu, Chu, Zang, Wei, Dong, Zhang, Liang, Xiong, Qiao, Lin, et~al.]{liu2024mmdu}
Ziyu Liu, Tao Chu, Yuhang Zang, Xilin Wei, Xiaoyi Dong, Pan Zhang, Zijian Liang, Yuanjun Xiong, Yu Qiao, Dahua Lin, et~al.
\newblock Mmdu: A multi-turn multi-image dialog understanding benchmark and instruction-tuning dataset for lvlms.
\newblock \emph{arXiv preprint arXiv:2406.11833}, 2024{\natexlab{d}}.

\bibitem[Lu et~al.(2024)Lu, Liu, Zhang, Wang, Dong, Liu, Sun, Ren, Li, Yang, et~al.]{lu2024deepseek}
Haoyu Lu, Wen Liu, Bo Zhang, Bingxuan Wang, Kai Dong, Bo Liu, Jingxiang Sun, Tongzheng Ren, Zhuoshu Li, Hao Yang, et~al.
\newblock Deepseek-vl: towards real-world vision-language understanding.
\newblock \emph{arxiv:2403.05525}, 2024.

\bibitem[Lu et~al.(2021)Lu, Gong, Jiang, Qiu, Huang, Liang, and Zhu]{lu2021inter}
Pan Lu, Ran Gong, Shibiao Jiang, Liang Qiu, Siyuan Huang, Xiaodan Liang, and Song-Chun Zhu.
\newblock Inter-gps: Interpretable geometry problem solving with formal language and symbolic reasoning.
\newblock \emph{arXiv preprint arXiv:2105.04165}, 2021.

\bibitem[Lu et~al.(2022{\natexlab{a}})Lu, Mishra, Xia, Qiu, Chang, Zhu, Tafjord, Clark, and Kalyan]{lu2022learn}
Pan Lu, Swaroop Mishra, Tanglin Xia, Liang Qiu, Kai-Wei Chang, Song-Chun Zhu, Oyvind Tafjord, Peter Clark, and Ashwin Kalyan.
\newblock Learn to explain: Multimodal reasoning via thought chains for science question answering.
\newblock In \emph{NeurIPS}, pages 2507--2521, 2022{\natexlab{a}}.

\bibitem[Lu et~al.(2022{\natexlab{b}})Lu, Qiu, Chang, Wu, Zhu, Rajpurohit, Clark, and Kalyan]{lu2022dynamic}
Pan Lu, Liang Qiu, Kai-Wei Chang, Ying~Nian Wu, Song-Chun Zhu, Tanmay Rajpurohit, Peter Clark, and Ashwin Kalyan.
\newblock Dynamic prompt learning via policy gradient for semi-structured mathematical reasoning.
\newblock \emph{arXiv preprint arXiv:2209.14610}, 2022{\natexlab{b}}.

\bibitem[Maaz et~al.(2024)Maaz, Rasheed, Khan, and Khan]{maaz2024videogpt+}
Muhammad Maaz, Hanoona Rasheed, Salman Khan, and Fahad Khan.
\newblock Videogpt+: Integrating image and video encoders for enhanced video understanding.
\newblock \emph{arXiv preprint arXiv:2406.09418}, 2024.

\bibitem[Mao et~al.(2017)Mao, Cheung, and She]{mao2017deepart}
Hui Mao, Ming Cheung, and James She.
\newblock Deepart: Learning joint representations of visual arts.
\newblock In \emph{Proceedings of the 25th ACM international conference on Multimedia}, pages 1183--1191, 2017.

\bibitem[Marino et~al.(2019)Marino, Rastegari, Farhadi, and Mottaghi]{marino2019ok}
Kenneth Marino, Mohammad Rastegari, Ali Farhadi, and Roozbeh Mottaghi.
\newblock Ok-vqa: A visual question answering benchmark requiring external knowledge.
\newblock In \emph{Proceedings of the IEEE/cvf conference on computer vision and pattern recognition}, pages 3195--3204, 2019.

\bibitem[Marti and Bunke(2002)]{marti2002iam}
U-V Marti and Horst Bunke.
\newblock The iam-database: an english sentence database for offline handwriting recognition.
\newblock \emph{International journal on document analysis and recognition}, 5:\penalty0 39--46, 2002.

\bibitem[Masry et~al.(2022)Masry, Long, Tan, Joty, and Hoque]{masry2022chartqa}
Ahmed Masry, Do~Xuan Long, Jia~Qing Tan, Shafiq Joty, and Enamul Hoque.
\newblock Chartqa: A benchmark for question answering about charts with visual and logical reasoning.
\newblock \emph{arxiv:2203.10244}, 2022.

\bibitem[Masry et~al.(2023)Masry, Kavehzadeh, Do, Hoque, and Joty]{masry2023unichart}
Ahmed Masry, Parsa Kavehzadeh, Xuan~Long Do, Enamul Hoque, and Shafiq Joty.
\newblock Unichart: A universal vision-language pretrained model for chart comprehension and reasoning.
\newblock \emph{arXiv preprint arXiv:2305.14761}, 2023.

\bibitem[Mathew et~al.(2021)Mathew, Karatzas, and Jawahar]{mathew2021docvqa}
Minesh Mathew, Dimosthenis Karatzas, and CV Jawahar.
\newblock Docvqa: A dataset for vqa on document images.
\newblock In \emph{WACV}, pages 2200--2209, 2021.

\bibitem[Mathew et~al.(2022)Mathew, Bagal, Tito, Karatzas, Valveny, and Jawahar]{mathew2022infographicvqa}
Minesh Mathew, Viraj Bagal, Rub{\`e}n Tito, Dimosthenis Karatzas, Ernest Valveny, and CV Jawahar.
\newblock Infographicvqa.
\newblock In \emph{WACV}, pages 1697--1706, 2022.

\bibitem[McKinzie et~al.(2024)McKinzie, Gan, Fauconnier, Dodge, Zhang, Dufter, Shah, Du, Peng, Weers, et~al.]{mckinzie2024mm1}
Brandon McKinzie, Zhe Gan, Jean-Philippe Fauconnier, Sam Dodge, Bowen Zhang, Philipp Dufter, Dhruti Shah, Xianzhi Du, Futang Peng, Floris Weers, et~al.
\newblock Mm1: Methods, analysis \& insights from multimodal llm pre-training.
\newblock \emph{arxiv:2403.09611}, 2024.

\bibitem[Methani et~al.(2020)Methani, Ganguly, Khapra, and Kumar]{methani2020plotqa}
Nitesh Methani, Pritha Ganguly, Mitesh~M Khapra, and Pratyush Kumar.
\newblock Plotqa: Reasoning over scientific plots.
\newblock In \emph{Proceedings of the IEEE/CVF Winter Conference on Applications of Computer Vision}, pages 1527--1536, 2020.

\bibitem[Park et~al.(2024)Park, Hessel, Chandu, Liang, Lu, West, Yu, Huang, Gao, Farhadi, et~al.]{park2024localized}
Jae~Sung Park, Jack Hessel, Khyathi Chandu, Paul~Pu Liang, Ximing Lu, Peter West, Youngjae Yu, Qiuyuan Huang, Jianfeng Gao, Ali Farhadi, et~al.
\newblock Localized symbolic knowledge distillation for visual commonsense models.
\newblock In \emph{NeurIPS}, 2024.

\bibitem[Peng et~al.(2023)Peng, Wang, Dong, Hao, Huang, Ma, and Wei]{peng2023kosmos}
Zhiliang Peng, Wenhui Wang, Li Dong, Yaru Hao, Shaohan Huang, Shuming Ma, and Furu Wei.
\newblock Kosmos-2: Grounding multimodal large language models to the world.
\newblock \emph{arXiv preprint arXiv:2306.14824}, 2023.

\bibitem[Pope et~al.(2023)Pope, Douglas, Chowdhery, Devlin, Bradbury, Heek, Xiao, Agrawal, and Dean]{pope2023efficiently}
Reiner Pope, Sholto Douglas, Aakanksha Chowdhery, Jacob Devlin, James Bradbury, Jonathan Heek, Kefan Xiao, Shivani Agrawal, and Jeff Dean.
\newblock Efficiently scaling transformer inference.
\newblock \emph{Proceedings of Machine Learning and Systems}, 5:\penalty0 606--624, 2023.

\bibitem[Rohrbach et~al.(2015)Rohrbach, Rohrbach, Tandon, and Schiele]{rohrbach2015dataset}
Anna Rohrbach, Marcus Rohrbach, Niket Tandon, and Bernt Schiele.
\newblock A dataset for movie description.
\newblock In \emph{Proceedings of the IEEE conference on computer vision and pattern recognition}, pages 3202--3212, 2015.

\bibitem[Schuhmann et~al.(2022{\natexlab{a}})Schuhmann, Beaumont, Vencu, Gordon, Wightman, Cherti, Coombes, Katta, Mullis, Wortsman, et~al.]{schuhmann2022laion}
Christoph Schuhmann, Romain Beaumont, Richard Vencu, Cade Gordon, Ross Wightman, Mehdi Cherti, Theo Coombes, Aarush Katta, Clayton Mullis, Mitchell Wortsman, et~al.
\newblock Laion-5b: An open large-scale dataset for training next generation image-text models.
\newblock In \emph{NeurIPS}, pages 25278--25294, 2022{\natexlab{a}}.

\bibitem[Schuhmann et~al.(2022{\natexlab{b}})Schuhmann, K{\"o}pf, Vencu, Coombes, and Beaumont]{schuhmann2022laion-coco}
Christoph Schuhmann, Andreas K{\"o}pf, Richard Vencu, Theo Coombes, and Romain Beaumont.
\newblock Laion coco: 600m synthetic captions from laion2b-en.
\newblock \url{https://laion.ai/blog/laion-coco/}, 2022{\natexlab{b}}.

\bibitem[Seo et~al.(2015)Seo, Hajishirzi, Farhadi, Etzioni, and Malcolm]{seo2015solving}
Minjoon Seo, Hannaneh Hajishirzi, Ali Farhadi, Oren Etzioni, and Clint Malcolm.
\newblock Solving geometry problems: Combining text and diagram interpretation.
\newblock In \emph{Proceedings of the 2015 conference on empirical methods in natural language processing}, pages 1466--1476, 2015.

\bibitem[Shah et~al.(2019)Shah, Mishra, Yadati, and Talukdar]{shah2019kvqa}
Sanket Shah, Anand Mishra, Naganand Yadati, and Partha~Pratim Talukdar.
\newblock Kvqa: Knowledge-aware visual question answering.
\newblock In \emph{Proceedings of the AAAI conference on artificial intelligence}, pages 8876--8884, 2019.

\bibitem[Shahroudy et~al.(2016)Shahroudy, Liu, Ng, and Wang]{shahroudy2016ntu}
Amir Shahroudy, Jun Liu, Tian-Tsong Ng, and Gang Wang.
\newblock Ntu rgb+d: A large scale dataset for 3d human activity analysis.
\newblock In \emph{Proceedings of the IEEE conference on computer vision and pattern recognition}, pages 1010--1019, 2016.

\bibitem[Shao et~al.(2019)Shao, Li, Zhang, Peng, Yu, Zhang, Li, and Sun]{shao2019objects365}
Shuai Shao, Zeming Li, Tianyuan Zhang, Chao Peng, Gang Yu, Xiangyu Zhang, Jing Li, and Jian Sun.
\newblock Objects365: A large-scale, high-quality dataset for object detection.
\newblock In \emph{Proceedings of the IEEE/CVF international conference on computer vision}, pages 8430--8439, 2019.

\bibitem[Singh et~al.(2019)Singh, Natarajan, Shah, Jiang, Chen, Batra, Parikh, and Rohrbach]{singh2019towards}
Amanpreet Singh, Vivek Natarajan, Meet Shah, Yu Jiang, Xinlei Chen, Dhruv Batra, Devi Parikh, and Marcus Rohrbach.
\newblock Towards vqa models that can read.
\newblock In \emph{CVPR}, pages 8317--8326, 2019.

\bibitem[Singh et~al.(2021)Singh, Pang, Toh, Huang, Galuba, and Hassner]{singh2021textocr}
Amanpreet Singh, Guan Pang, Mandy Toh, Jing Huang, Wojciech Galuba, and Tal Hassner.
\newblock Textocr: Towards large-scale end-to-end reasoning for arbitrary-shaped scene text.
\newblock In \emph{Proceedings of the IEEE/CVF conference on computer vision and pattern recognition}, pages 8802--8812, 2021.

\bibitem[Sun et~al.(2024)Sun, Cui, Zhang, Zhang, Yu, Wang, Rao, Liu, Huang, and Wang]{sun2024generative}
Quan Sun, Yufeng Cui, Xiaosong Zhang, Fan Zhang, Qiying Yu, Yueze Wang, Yongming Rao, Jingjing Liu, Tiejun Huang, and Xinlong Wang.
\newblock Generative multimodal models are in-context learners.
\newblock In \emph{CVPR}, pages 14398--14409, 2024.

\bibitem[Sun et~al.(2019)Sun, Ni, Chng, Liu, Luo, Ng, Han, Ding, Liu, Karatzas, et~al.]{sun2019icdar}
Yipeng Sun, Zihan Ni, Chee-Kheng Chng, Yuliang Liu, Canjie Luo, Chun~Chet Ng, Junyu Han, Errui Ding, Jingtuo Liu, Dimosthenis Karatzas, et~al.
\newblock Icdar 2019 competition on large-scale street view text with partial labeling-rrc-lsvt.
\newblock In \emph{2019 International Conference on Document Analysis and Recognition (ICDAR)}, pages 1557--1562. IEEE, 2019.

\bibitem[Tao et~al.(2024)Tao, Zhu, Su, Lu, Tian, Luo, Huang, Li, Qiao, Zhou, et~al.]{tao2024learning}
Chenxin Tao, Xizhou Zhu, Shiqian Su, Lewei Lu, Changyao Tian, Xuan Luo, Gao Huang, Hongsheng Li, Yu Qiao, Jie Zhou, et~al.
\newblock Learning 1d causal visual representation with de-focus attention networks.
\newblock \emph{arxiv:2406.04342}, 2024.

\bibitem[Team(2024)]{team2024chameleon}
Chameleon Team.
\newblock Chameleon: Mixed-modal early-fusion foundation models.
\newblock \emph{arxiv:2405.09818}, 2024.

\bibitem[Team et~al.(2023)Team, Anil, Borgeaud, Alayrac, Yu, Soricut, Schalkwyk, Dai, Hauth, Millican, et~al.]{team2023gemini}
Gemini Team, Rohan Anil, Sebastian Borgeaud, Jean-Baptiste Alayrac, Jiahui Yu, Radu Soricut, Johan Schalkwyk, Andrew~M Dai, Anja Hauth, Katie Millican, et~al.
\newblock Gemini: a family of highly capable multimodal models.
\newblock \emph{arxiv:2312.11805}, 2023.

\bibitem[Touvron et~al.(2023)Touvron, Lavril, Izacard, Martinet, Lachaux, Lacroix, Rozi{\`e}re, Goyal, Hambro, Azhar, et~al.]{touvron2023llama}
Hugo Touvron, Thibaut Lavril, Gautier Izacard, Xavier Martinet, Marie-Anne Lachaux, Timoth{\'e}e Lacroix, Baptiste Rozi{\`e}re, Naman Goyal, Eric Hambro, Faisal Azhar, et~al.
\newblock Llama: Open and efficient foundation language models.
\newblock \emph{arxiv:2302.13971}, 2023.

\bibitem[Veit et~al.(2016)Veit, Matera, Neumann, Matas, and Belongie]{veit2016coco}
Andreas Veit, Tomas Matera, Lukas Neumann, Jiri Matas, and Serge Belongie.
\newblock Coco-text: Dataset and benchmark for text detection and recognition in natural images.
\newblock \emph{arXiv preprint arXiv:1601.07140}, 2016.

\bibitem[Wang et~al.(2023{\natexlab{a}})Wang, Zhang, Chu, Cao, Zhou, Wu, Wang, He, and Lin]{wang2023v3det}
Jiaqi Wang, Pan Zhang, Tao Chu, Yuhang Cao, Yujie Zhou, Tong Wu, Bin Wang, Conghui He, and Dahua Lin.
\newblock V3det: Vast vocabulary visual detection dataset.
\newblock In \emph{Proceedings of the IEEE/CVF International Conference on Computer Vision}, pages 19844--19854, 2023{\natexlab{a}}.

\bibitem[Wang et~al.(2024{\natexlab{a}})Wang, Bai, Tan, Wang, Fan, Bai, Chen, Liu, Wang, Ge, et~al.]{wang2024qwen2}
Peng Wang, Shuai Bai, Sinan Tan, Shijie Wang, Zhihao Fan, Jinze Bai, Keqin Chen, Xuejing Liu, Jialin Wang, Wenbin Ge, et~al.
\newblock Qwen2-vl: Enhancing vision-language model's perception of the world at any resolution.
\newblock \emph{arxiv:2409.12191}, 2024{\natexlab{a}}.

\bibitem[Wang et~al.(2022)Wang, Zhou, Zeng, and Zhang]{wang2022efficientvlm}
Tiannan Wang, Wangchunshu Zhou, Yan Zeng, and Xinsong Zhang.
\newblock Efficientvlm: Fast and accurate vision-language models via knowledge distillation and modal-adaptive pruning.
\newblock \emph{arxiv:2210.07795}, 2022.

\bibitem[Wang et~al.(2023{\natexlab{b}})Wang, Shi, Li, Wang, Huang, Xing, Chen, Li, Zhu, Cao, et~al.]{wang2023all}
Weiyun Wang, Min Shi, Qingyun Li, Wenhai Wang, Zhenhang Huang, Linjie Xing, Zhe Chen, Hao Li, Xizhou Zhu, Zhiguo Cao, et~al.
\newblock The all-seeing project: Towards panoptic visual recognition and understanding of the open world.
\newblock \emph{arXiv preprint arXiv:2308.01907}, 2023{\natexlab{b}}.

\bibitem[Wang et~al.(2024{\natexlab{b}})Wang, Chen, Chen, Wu, Zhu, Zeng, Luo, Lu, Zhou, Qiao, et~al.]{wang2024visionllm}
Wenhai Wang, Zhe Chen, Xiaokang Chen, Jiannan Wu, Xizhou Zhu, Gang Zeng, Ping Luo, Tong Lu, Jie Zhou, Yu Qiao, et~al.
\newblock Visionllm: Large language model is also an open-ended decoder for vision-centric tasks.
\newblock In \emph{NeurIPS}, 2024{\natexlab{b}}.

\bibitem[Wang et~al.(2020)Wang, Liu, Shen, Ng, Luo, Jin, Chan, Hengel, and Wang]{wang2020general}
Xinyu Wang, Yuliang Liu, Chunhua Shen, Chun~Chet Ng, Canjie Luo, Lianwen Jin, Chee~Seng Chan, Anton van~den Hengel, and Liangwei Wang.
\newblock On the general value of evidence, and bilingual scene-text visual question answering.
\newblock In \emph{Proceedings of the IEEE/CVF Conference on Computer Vision and Pattern Recognition}, pages 10126--10135, 2020.

\bibitem[Wang et~al.(2024{\natexlab{c}})Wang, Zhang, Luo, Sun, Cui, Wang, Zhang, Wang, Li, Yu, et~al.]{wang2024emu3}
Xinlong Wang, Xiaosong Zhang, Zhengxiong Luo, Quan Sun, Yufeng Cui, Jinsheng Wang, Fan Zhang, Yueze Wang, Zhen Li, Qiying Yu, et~al.
\newblock Emu3: Next-token prediction is all you need.
\newblock \emph{arxiv:2409.18869}, 2024{\natexlab{c}}.

\bibitem[Wang et~al.(2024{\natexlab{d}})Wang, Zhou, Liu, Lu, Xu, He, Yoon, Lu, Bertasius, Bansal, et~al.]{wang2024mementos}
Xiyao Wang, Yuhang Zhou, Xiaoyu Liu, Hongjin Lu, Yuancheng Xu, Feihong He, Jaehong Yoon, Taixi Lu, Gedas Bertasius, Mohit Bansal, et~al.
\newblock Mementos: A comprehensive benchmark for multimodal large language model reasoning over image sequences.
\newblock \emph{arXiv preprint arXiv:2401.10529}, 2024{\natexlab{d}}.

\bibitem[Wu et~al.(2024)Wu, Yu, Chen, Tenenbaum, and Gan]{wu2024star}
Bo Wu, Shoubin Yu, Zhenfang Chen, Joshua~B Tenenbaum, and Chuang Gan.
\newblock Star: A benchmark for situated reasoning in real-world videos.
\newblock \emph{arXiv preprint arXiv:2405.09711}, 2024.

\bibitem[Wu(2023)]{pmc2023case}
Chaoyi Wu.
\newblock Pmc-casereport.
\newblock \url{https://huggingface.co/datasets/chaoyi-wu/PMC-CaseReport}, 2023.

\bibitem[X.ai(2024)]{grok}
X.ai.
\newblock Grok-1.5 vision preview.
\newblock \url{https://x.ai/blog/grok-1.5v}, 2024.

\bibitem[Xu et~al.(2024)Xu, Yao, Guo, Cui, Ni, Ge, Chua, Liu, Sun, and Huang]{xu2024llava}
Ruyi Xu, Yuan Yao, Zonghao Guo, Junbo Cui, Zanlin Ni, Chunjiang Ge, Tat-Seng Chua, Zhiyuan Liu, Maosong Sun, and Gao Huang.
\newblock Llava-uhd: an lmm perceiving any aspect ratio and high-resolution images.
\newblock \emph{arXiv:2403.11703}, 2024.

\bibitem[Yang et~al.(2023)Yang, Li, Lin, Wang, Lin, Liu, and Wang]{yang2023dawn}
Zhengyuan Yang, Linjie Li, Kevin Lin, Jianfeng Wang, Chung-Ching Lin, Zicheng Liu, and Lijuan Wang.
\newblock The dawn of lmms: Preliminary explorations with gpt-4v (ision).
\newblock \emph{arxiv:2309.17421}, 9\penalty0 (1):\penalty0 1, 2023.

\bibitem[Yao et~al.(2024)Yao, Yu, Zhang, Wang, Cui, Zhu, Cai, Li, Zhao, He, et~al.]{yao2024minicpm}
Yuan Yao, Tianyu Yu, Ao Zhang, Chongyi Wang, Junbo Cui, Hongji Zhu, Tianchi Cai, Haoyu Li, Weilin Zhao, Zhihui He, et~al.
\newblock Minicpm-v: A gpt-4v level mllm on your phone.
\newblock \emph{arxiv:2408.01800}, 2024.

\bibitem[Yi et~al.(2019)Yi, Gan, Li, Kohli, Wu, Torralba, and Tenenbaum]{yi2019clevrer}
Kexin Yi, Chuang Gan, Yunzhu Li, Pushmeet Kohli, Jiajun Wu, Antonio Torralba, and Joshua~B Tenenbaum.
\newblock Clevrer: Collision events for video representation and reasoning.
\newblock \emph{arXiv preprint arXiv:1910.01442}, 2019.

\bibitem[Yu et~al.(2023{\natexlab{a}})Yu, Jiang, Shi, Yu, Liu, Zhang, Kwok, Li, Weller, and Liu]{yu2023metamath}
Longhui Yu, Weisen Jiang, Han Shi, Jincheng Yu, Zhengying Liu, Yu Zhang, James~T Kwok, Zhenguo Li, Adrian Weller, and Weiyang Liu.
\newblock Metamath: Bootstrap your own mathematical questions for large language models.
\newblock \emph{arXiv preprint arXiv:2309.12284}, 2023{\natexlab{a}}.

\bibitem[Yu et~al.(2023{\natexlab{b}})Yu, Yang, Li, Wang, Lin, Liu, Wang, and Wang]{yu2023mmvet}
Weihao Yu, Zhengyuan Yang, Linjie Li, Jianfeng Wang, Kevin Lin, Zicheng Liu, Xinchao Wang, and Lijuan Wang.
\newblock Mm-vet: Evaluating large multimodal models for integrated capabilities.
\newblock \emph{arxiv:2308.02490}, 2023{\natexlab{b}}.

\bibitem[Yuan et~al.(2019)Yuan, Zhu, Xu, Li, Mu, and Hu]{yuan2019ctw}
Tai{-}Ling Yuan, Zhe Zhu, Kun Xu, Cheng{-}Jun Li, Tai{-}Jiang Mu, and Shi{-}Min Hu.
\newblock A large chinese text dataset in the wild.
\newblock \emph{Journal of Computer Science and Technology}, 34\penalty0 (3):\penalty0 509--521, 2019.

\bibitem[Yuan et~al.(2022)Yuan, Liu, Dikubab, Liu, Ji, Wu, and Bai]{yuan2022syntax}
Ye Yuan, Xiao Liu, Wondimu Dikubab, Hui Liu, Zhilong Ji, Zhongqin Wu, and Xiang Bai.
\newblock Syntax-aware network for handwritten mathematical expression recognition.
\newblock \emph{arXiv preprint arXiv:2203.01601}, 2022.

\bibitem[Yue et~al.(2024)Yue, Ni, Zhang, Zheng, Liu, Zhang, Stevens, Jiang, Ren, Sun, et~al.]{yue2024mmmu}
Xiang Yue, Yuansheng Ni, Kai Zhang, Tianyu Zheng, Ruoqi Liu, Ge Zhang, Samuel Stevens, Dongfu Jiang, Weiming Ren, Yuxuan Sun, et~al.
\newblock Mmmu: A massive multi-discipline multimodal understanding and reasoning benchmark for expert agi.
\newblock In \emph{CVPR}, pages 9556--9567, 2024.

\bibitem[Zhang et~al.(2019)Zhang, Zhou, Jiang, Song, Li, Zhou, Wang, Wang, Liao, Yang, et~al.]{zhang2019icdar}
Rui Zhang, Yongsheng Zhou, Qianyi Jiang, Qi Song, Nan Li, Kai Zhou, Lei Wang, Dong Wang, Minghui Liao, Mingkun Yang, et~al.
\newblock Icdar 2019 robust reading challenge on reading chinese text on signboard.
\newblock In \emph{2019 international conference on document analysis and recognition (ICDAR)}, pages 1577--1581. IEEE, 2019.

\bibitem[Zhang et~al.(2023{\natexlab{a}})Zhang, Wu, Zhao, Lin, Zhang, Wang, and Xie]{zhang2023pmc}
Xiaoman Zhang, Chaoyi Wu, Ziheng Zhao, Weixiong Lin, Ya Zhang, Yanfeng Wang, and Weidi Xie.
\newblock Pmc-vqa: Visual instruction tuning for medical visual question answering.
\newblock \emph{arXiv preprint arXiv:2305.10415}, 2023{\natexlab{a}}.

\bibitem[Zhang et~al.(2023{\natexlab{b}})Zhang, Zhang, Gu, Zhou, Lipka, Yang, and Sun]{zhang2023llavar}
Yanzhe Zhang, Ruiyi Zhang, Jiuxiang Gu, Yufan Zhou, Nedim Lipka, Diyi Yang, and Tong Sun.
\newblock Llavar: Enhanced visual instruction tuning for text-rich image understanding.
\newblock \emph{arXiv preprint arXiv:2306.17107}, 2023{\natexlab{b}}.

\bibitem[Zhao et~al.(2023)Zhao, Zhou, Li, Tang, Wang, Hou, Min, Zhang, Zhang, Dong, et~al.]{zhao2023survey}
Wayne~Xin Zhao, Kun Zhou, Junyi Li, Tianyi Tang, Xiaolei Wang, Yupeng Hou, Yingqian Min, Beichen Zhang, Junjie Zhang, Zican Dong, et~al.
\newblock A survey of large language models.
\newblock \emph{arxiv:2303.18223}, 2023.

\bibitem[Zheng et~al.(2021)Zheng, Burdick, Popa, Zhong, and Wang]{zheng2021global}
Xinyi Zheng, Douglas Burdick, Lucian Popa, Xu Zhong, and Nancy Xin~Ru Wang.
\newblock Global table extractor (gte): A framework for joint table identification and cell structure recognition using visual context.
\newblock In \emph{Proceedings of the IEEE/CVF winter conference on applications of computer vision}, pages 697--706, 2021.

\bibitem[Zhou et~al.(2024)Zhou, Yu, Babu, Tirumala, Yasunaga, Shamis, Kahn, Ma, Zettlemoyer, and Levy]{zhou2024transfusion}
Chunting Zhou, Lili Yu, Arun Babu, Kushal Tirumala, Michihiro Yasunaga, Leonid Shamis, Jacob Kahn, Xuezhe Ma, Luke Zettlemoyer, and Omer Levy.
\newblock Transfusion: Predict the next token and diffuse images with one multi-modal model.
\newblock \emph{arxiv:2408.11039}, 2024.

\bibitem[Zhu et~al.(2023)Zhu, Chen, Shen, Li, and Elhoseiny]{zhu2023minigpt}
Deyao Zhu, Jun Chen, Xiaoqian Shen, Xiang Li, and Mohamed Elhoseiny.
\newblock Minigpt-4: Enhancing vision-language understanding with advanced large language models.
\newblock \emph{arxiv:2304.10592}, 2023.

\end{thebibliography}
}

\end{document}